\documentclass{article}

\usepackage{PRIMEarxiv}

\usepackage[utf8]{inputenc} 
\usepackage[T1]{fontenc}    
\usepackage{hyperref}       
\usepackage{url}            
\usepackage{booktabs}       
\usepackage{amsfonts}       
\usepackage{nicefrac}       
\usepackage{microtype}      
\usepackage{lipsum}
\usepackage{fancyhdr}       
\usepackage{graphicx}       

\usepackage{adjustbox}
\usepackage{amsmath}
\usepackage{array}
\usepackage{booktabs}
\usepackage{longtable}
\usepackage{caption}
\usepackage{multirow}
\usepackage{multicol}
\usepackage{rotating}
\usepackage{tabularx}
\usepackage{url}
\usepackage[table,xcdraw]{xcolor}
\usepackage{hyperref} 
\usepackage{balance}

\title{
    A Systematic Review of ECG Arrhythmia Classification: Adherence to Standards, Fair Evaluation, and Embedded Feasibility
}

\author{
    Guilherme A. L. Silva, Pedro H. L. Silva, Gladston J. P. Moreira, Vander L. S. Freitas, \\ \textbf{Jadson C. Gertrudes and Eduardo J. S. Luz}\\
    Computing Department, Federal University of Ouro Preto\\
    Ouro Preto, Brazil\\
    \texttt{guilherme.lopes@aluno.ufop.edu.br}\\ \texttt{\{silvap,gladston,vander.freitas,jadson.castro,eduluz\}@ufop.edu.br} \\
}

\begin{document}
\maketitle

\begin{abstract}
    The classification of electrocardiogram (ECG) signals is crucial for early detection of arrhythmias and other cardiac conditions. However, despite advances in machine learning-based approaches, a significant portion of the literature does not adhere to key standardization protocols, leading to inconsistencies in performance evaluation and real-world applicability. 
    Additionally, many studies overlook hardware constraints essential for real-world deployment, such as in pacemakers, Holter monitors, wearable ECG patches, and implantable loop recorders. Since the practical impact of these models is often tied to their feasibility in resource-constrained embedded systems, ensuring efficient deployment in such devices is critical for practical, remote, or continuous cardiac monitoring. Given this context, our review focuses on studies that address the challenges of deploying ECG classification models in embedded environments and systematically reviews ECG classification research published between 2017 and 2024, focusing on studies that follow the E3C  (Embedded, Clinical, and Comparative Criteria), which encompass the implementation of the inter-patient paradigm, compliance with Association for the Advancement of Medical Instrumentation (AAMI) recommendations, and evaluation of model feasibility for deployment in resource-constrained environments. Our findings highlight that while many studies achieve high classification accuracy, only a minority properly account for real-world constraints such as patient-independent data partitioning and hardware limitations. Furthermore, we identify the state-of-the-art methods that meet the E3C.
    Through a detailed comparative analysis, we assess the trade-offs between accuracy, inference time, energy consumption, and memory usage across different classification models. Finally, we provide recommendations for standardized performance reporting to ensure fair comparisons and practical applicability of ECG classification models. By addressing these critical gaps, this study aims to guide future research towards more robust, transparent, and clinically viable ECG classification systems.
\end{abstract}


\section{Introduction}
\label{sec:introduction}

    According to the World Health Organization (WHO), cardiovascular diseases remain the leading cause of death globally, responsible for an estimated 17.9 million fatalities each year \cite{whoheart}. Approximately 75\% of these cases occur in low- and middle-income countries, highlighting the need for accessible diagnostic tools in diverse healthcare settings. The Electrocardiogram (ECG), due to its non-invasive nature and simplicity, is widely used for diagnosing heart diseases \cite{cohen_biomedical_1986}. The early detection of arrhythmias is essential, as it can significantly improve treatment outcomes and help in the diagnosis of potentially life-threatening conditions.
    
    ECGs can reveal a variety of heart diseases \cite{luz_ecg-based_2016}, with arrhythmias — abnormal or irregular heartbeats — being a critical indicator of cardiovascular health. Some types of arrhythmias are harmless, while others can lead to severe health diseases, including heart attacks and sudden death. Arrhythmias are generally categorized into morphological arrhythmias, characterized by isolated irregular heartbeats, and rhythmic arrhythmias, involving sequences of irregular beats. Rhythmic arrhythmias, in particular, require urgent identification and classification due to their association with serious cardiac conditions.
    
    Traditionally, arrhythmia diagnosis involves manual beat-by-beat inspection of ECG recordings, a time-consuming and error-prone process for physicians, especially in the case of long-term monitoring. Real-time classification, such as in bedside or wearable monitoring, presents additional challenges, particularly for less experienced physicians. As a solution, machine learning models \cite{meng2022enhancing, alamatsaz2024lightweight, chen2024novel, le2023lightx3ecg} have been developed to automate the classification of ECG patterns, aiming to reduce diagnostic times and improve accuracy in arrhythmia detection.
    
    Over the years, numerous studies have explored automatic arrhythmia detection and classification using machine learning \cite{amirshahi2019ecg, anand2024enhanced, cheng2022efficient, feyisa2022lightweight}. However, few studies in the literature follow the recommendations of the Association for the Advancement of Medical Instrumentation (AAMI), despite their critical role in establishing standardized evaluation protocols. The ANSI/AAMI/ISO EC57 standard provides guidelines for assessing arrhythmia classification algorithms, recommending the use of certified databases, with the MIT-BIH database \cite{moody_impact_2001} being one of the most widely adopted resources.
    
    Despite these guidelines, AAMI does not specify data partitioning protocols for training and testing, which has led to inconsistent results across studies. Many models show artificial high accuracy by including beats from the same patient in both training and testing sets. This practice compromises generalization and does not reflect real-world application. To address this problem, De Chazal \textit{et al.} \cite{de_chazal_automatic_2004} proposed an inter-patient paradigm, in which training and testing sets are strictly separated by patients, ensuring that signals from a single individual do not appear in both sets. Under this protocol, performance metrics of algorithms tend to decrease, highlighting the complexity of arrhythmia classification when faced with patient-independent data — a condition much closer to practical deployment \cite{luz_how_2011}. 
    
    Implementing both AAMI recommendations and the inter-patient paradigm offers a more reliable assessment, but few studies have adopted this combined approach. Those that do, such as \cite{farag2023tiny,yan2021energy,mao2022ultra,xing2022accurate}, present more accurate insights into real-world performance. However, the high computational complexity of many proposed models, particularly in deep learning approaches, raises additional concerns. Most studies prioritize accuracy without considering the feasibility of deploying these models on low-power embedded systems, an aspect that is critical for practical, remote, or continuous cardiac monitoring. The need for low-power, efficient models is paramount for deployment in widely-used cardiac devices with constrained computational power, such as Holter monitors, wearable ECG patches, and implantable loop recorders.
    
    Deploying neural networks on these low-power devices for arrhythmia detection requires an emphasis on model optimization techniques, such as quantization  \cite{ran2022homecare,ribeiro2022ecg}, pruning \cite{rawal2023hardware,le2023lightx3ecg, ran2022homecare,lu2022efficient}, and Knowledge distillation \cite{saadatnejad2019lstm,meng2022enhancing,alamatsaz2024lightweight,chen2024novel,le2023lightx3ecg,ribeiro2022ecg,tesfai2022lightweight,xing2022accurate,sakib2021proof,jeon2020lightweight,feyisa2022lightweight}. Quantization, for example, reduces the bit-width of network parameters and activations, significantly reducing memory usage and computational demand without severely impacting accuracy. Pruning techniques remove redundant connections in the network, creating a more compact and efficient model. Knowledge distillation, where a large, accurate model ``teaches'' a smaller model, can also be employed to create lightweight networks suited for real-time operation on embedded platforms. In addition to traditional microcontrollers, Field-Programmable Gate Arrays (FPGAs) are increasingly considered for arrhythmia detection in wearable and portable devices \cite{rawal2023hardware,chu2022neuromorphic,ran2022homecare,xing2022accurate,tsoutsouras2017exploration,lu2022efficient,zhang2023configurable,cheng2022efficient,lu2021efficient}, as they offer customizable hardware configurations and enable parallel processing for improved performance. FPGAs can support optimized neural networks with low latency and power consumption.

    Many studies focus on optimizing inference performance \cite{farag2023tiny,xing2022accurate,raj2018personalized}, but they typically rely on externally trained models with fixed parameters, limiting their ability to adapt to diverse patient populations and dynamic physiological conditions. In contrast, on-device learning \cite{mao2022ultra,zhang2023configurable,cheng2022efficient} enables continuous model adaptation to patient-specific variations without requiring external retraining. This capability is particularly valuable for wearable and edge computing applications, where real-time learning ensures more robust and personalized ECG monitoring \cite{mao2022ultra}. By allowing models to refine predictions based on individual cardiac patterns \cite{dhar2021survey}, on-device learning enhances diagnostic accuracy and long-term performance.

    In this context, this study systematically reviews the literature from 2017 to 2024, expanding upon the survey conducted by Luz et al. \cite{luz_ecg-based_2016}, which did not explore embedded implementations since this area was still in its early stages at the time. The primary focus of our review is to analyze studies that evaluate the feasibility of deploying ECG classification models in embedded systems or low-cost hardware. We specifically examine the most cited studies in the literature, as well as the most relevant works from 2022, 2023, and 2024, while prioritizing those that investigate real-world deployment scenarios.

    Within this selected subset, we identify articles that comply with AAMI recommendations, implement the inter-patient paradigm as defined by De Chazal et al. \cite{de_chazal_automatic_2004}, and incorporate embedded solutions. We refer to this combination of criteria as E3C (Embedded, Clinical, and Comparative Criteria). Our goal is to assess whether these studies realistically simulate deployment conditions for ECG classification. Despite the advancements in deep learning, we observe that many models are still evaluated using intra-patient methods, leading to inflated performance metrics and limited generalization. Moreover, a significant portion of the literature lacks an assessment of model feasibility for embedded deployment, restricting their practical applicability in resource-constrained environments. By addressing these gaps, we aim to provide insights into the state of arrhythmia classification in real-world, low-resource settings and propose guidelines for standardized and fair performance reporting.

    The remainder of this article is structured as follows: 
    Section \ref{sec:aami} examines the AAMI standards and their relevance to ECG classification. Section \ref{sec:chazal} explores the inter-patient paradigm, focusing on the one proposed by De Chazal \textit{et al.} \cite{de_chazal_automatic_2004} and its implications for model evaluation. Section \ref{sec:hardware} discusses techniques for implementing ECG models in low-power environments, focusing on optimization strategies for embedded systems. Section \ref{sec:methodology} outlines the methodology used to select and analyze the articles included in this systematic review. Section \ref{sec:resultados} presents the findings of the review, highlighting the most relevant studies for low-power deployment. Finally, Section \ref{sec:conclusions} summarizes the key insights of this systematic review and discusses future research directions.

\section{Methodology}
\label{sec:methodology}

    The methodology for this systematic review involved a rigorous and structured approach to systematically identify and evaluate studies related to arrhythmia classification. The focus was on selecting the most cited articles and those that specifically addressed the deployment of ECG classification models, particularly in low-cost environments. Our goal was to analyze works that explored the feasibility of embedding models in resource-constrained hardware, such as microcontrollers and FPGAs, which are essential for real-time cardiac monitoring in scenarios with limited computational resources.
    These systems are critical for practical, real-time cardiac monitoring, particularly in devices where computational resources are limited.
    
    To gather the initial dataset of 1,427 articles, a structured search was conducted on the Web of Science database, with the following query: `TS=(``ECG classification'' OR ``arrhythmia detection'')', where `TS' specifies a topic search. It searches for articles containing the specified terms — ``ECG classification'' or ``arrhythmia detection'' — in the title, abstract, or keywords, ensuring the inclusion of studies directly relevant to this work's focus on arrhythmia detection via ECG. The search was further refined by applying two filters: journals with an impact factor greater than 1 and the publication period from 2017 to 2024. This approach ensured a broad yet relevant collection of high-quality literature, forming the foundation for a detailed and systematic review. The list of articles with citation counts was generated on August 27, 2024, ensuring the inclusion of the most recent and impactful studies.
    
\begin{figure*}[!t]
        \centerline{\includegraphics[width=\textwidth]{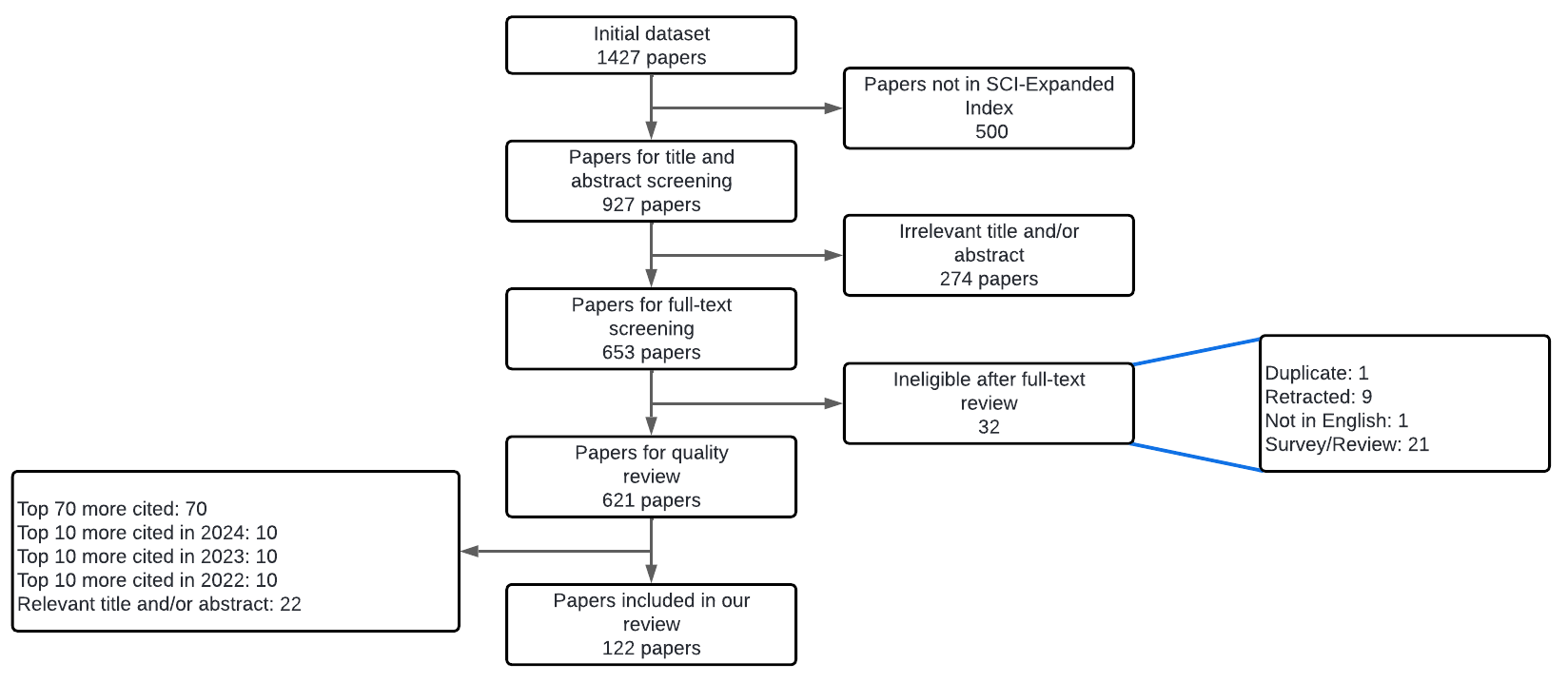}}
        \caption{Flowchart of the article selection process. The systematic review followed a multi-stage selection process to refine the dataset and ensure the inclusion of relevant, high-quality studies. First, the Science Citation Index Expanded (SCI-EXPANDED) was chosen within Web of Science, reducing the dataset to 927 articles by excluding 500 papers from other indexes. Next, a title and abstract screening removed 274 irrelevant studies, leaving 653 articles. A quality review further excluded 32 articles due to duplication, non-English content, unsuitable formats, or retraction. The final selection was based on two criteria: citation-based filtering, which identified 100 highly cited articles, and feasibility-based filtering, which added 22 studies focused on embedded implementation. This process resulted in a final dataset of 122 articles for analysis.}
        \label{fig:methodology}
    \end{figure*}

    The process of refining this dataset was carried out in multiple stages, as depicted in Figure \ref{fig:methodology}.
    
    The first stage of the selection process involved refining the dataset by selecting an index within Web of Science that prioritizes high-quality, peer-reviewed research. Given the technical nature of this study, the Science Citation Index Expanded (SCI-EXPANDED) was chosen as it provides the most comprehensive coverage of journals in engineering, medical sciences, and applied computing, ensuring that the selected studies are widely recognized and influential within their respective domains. Other available indexes, such as the Conference Proceedings Citation Index – Science (CPCI-S), focus primarily on conference papers, which often present preliminary findings rather than fully validated research. The Emerging Sources Citation Index (ESCI) includes journals still undergoing evaluation for inclusion in more established citation indexes, making their long-term impact less certain. The Social Sciences Citation Index (SSCI) and the Conference Proceedings Citation Index – Social Science \& Humanities (CPCI-SSH) primarily cover fields outside the study’s scope, such as education, policy, and humanities research. By filtering the dataset using SCI-EXPANDED, the selection was refined to 927 articles, excluding 500 papers that did not meet the criteria for high-impact, rigorously peer-reviewed research.
    
    In the next stage, the remaining 927 articles underwent a title and abstract screening. This phase involved a detailed assessment of each article’s title and abstract to determine its relevance to the systematic review’s focus on arrhythmia or ECG classification methods. Studies that did not involve arrhythmia or ECG classification were excluded, leading to the removal of 274 articles with ineligible content. Consequently, 653 articles proceeded to the quality review stage.
    
    The quality review, as represented in Figure \ref{fig:methodology}, involved filtering out articles that were not in English, retracted, duplicated, or classified as surveys or review papers. This step ensures the selection of high-quality studies that present ECG classification methods, focusing on methodological robustness and relevance to the field.

    During this phase, a total of 32 articles were removed for specific reasons, including duplicate publications (1), non-English language content (1), and unsuitable formats (21). Importantly, 9 retracted articles were also excluded from the final selection. A retracted article refers to a study that has been formally withdrawn from the academic record, often due to issues such as errors, falsified data, ethical concerns, or instances of plagiarism. Retracted articles are excluded from this systematic review because their findings are no longer deemed reliable by the scientific community. Including such studies could compromise the integrity of this review and introduce biases or inaccuracies. By excluding these retracted articles, this work ensures that only valid, peer-reviewed data contributes to the analysis, thereby upholding the quality and reliability of the findings.

    The final selection process for this systematic review was guided by two key criteria, resulting in the inclusion of 122 articles for analysis. The first criterion aimed to identify the most cited articles among the remaining 621 papers, ensuring that the review included foundational and widely recognized studies in the field. This stage included 100 articles. Initially, the 70 most cited articles were selected. However, to account for recent works that had not yet accumulated high citation counts but could still be highly relevant, the selection was expanded to include the 10 most cited articles from 2024, 2023, and 2022.
    
    The second criterion focused on identifying studies that explored the feasibility of deploying models in low-power embedded systems, an aspect crucial to real-world implementation. This analysis was conducted after the citation-based filtering to ensure that, beyond the most influential studies, the review also captured relevant works that, despite not being highly cited, provided meaningful contributions to embedded feasibility. Specifically, the selection process sought studies that employed lightweight neural networks, quantization or pruning techniques, FPGA implementations, or the development of dedicated ECG processors, as these techniques indicate an explicit effort to assess real-world deployment constraints. This additional filtering led to the inclusion of 22 more articles, further expanding the scope of the review by integrating research focused on the practical implementation of ECG classification in resource-constrained environments.
    
    Each of the final 122 articles was analyzed with a focus on methodology, particularly assessing compliance with E3C—our designation for studies that adhere to AAMI standards, implement the inter-patient paradigm, and evaluate computational efficiency for embedded deployment.
    Articles were classified into those that adhered to both AAMI and inter-patient protocols, ensuring a fair and realistic assessment of model performance, and those that addressed feasibility for embedded systems. Emphasis was placed on studies that proposed lightweight, efficient models capable of real-time operation on low-power hardware, such as FPGAs and microcontrollers, which are commonly used in portable and wearable cardiac devices like Holter monitors and wearable ECG patches. This attention to low-power compatibility is crucial for real-world applications, as many existing models in the literature, though accurate, are computationally intensive and not viable for deployment in constrained environments.

\section{AAMI Recommendations in Arrhythmia Classification}
\label{sec:aami}

    The AAMI recommendations are paramount to understanding the importance of standardization and rigorous evaluation in arrhythmia classification. Although many studies focus on developing advanced arrhythmia detection algorithms, few follow the guidelines established by AAMI, which compromises result consistency and comparability. This section discusses AAMI standards, adherence to these standards in current literature, and limitations observed in the implementation of these recommendations.
    
    \subsection{AAMI Standards}
    
        The ANSI/AAMI EC57:1998/(R)2008 standard \cite{aami:2008} establishes directives for algorithms aimed at detecting cardiac rhythm disturbances, providing guidance for robust testing and fair reporting of results.

        Among the databases commonly referenced in the literature, the MIT-BIH Arrhythmia Database (described in detail in Section \ref{sec:mit}) offers the most extensive diversity of arrhythmias and beat types. Due to this diversity, it is one of the most widely used databases in published studies, including this systematic review. Additionally, the MIT-BIH database was the first publicly available standardized set of test material designed specifically for evaluating arrhythmia detectors \cite{moody_impact_2001}.
        
\begin{figure}[!htb]
            \centering
            \includegraphics[width=0.5\linewidth]{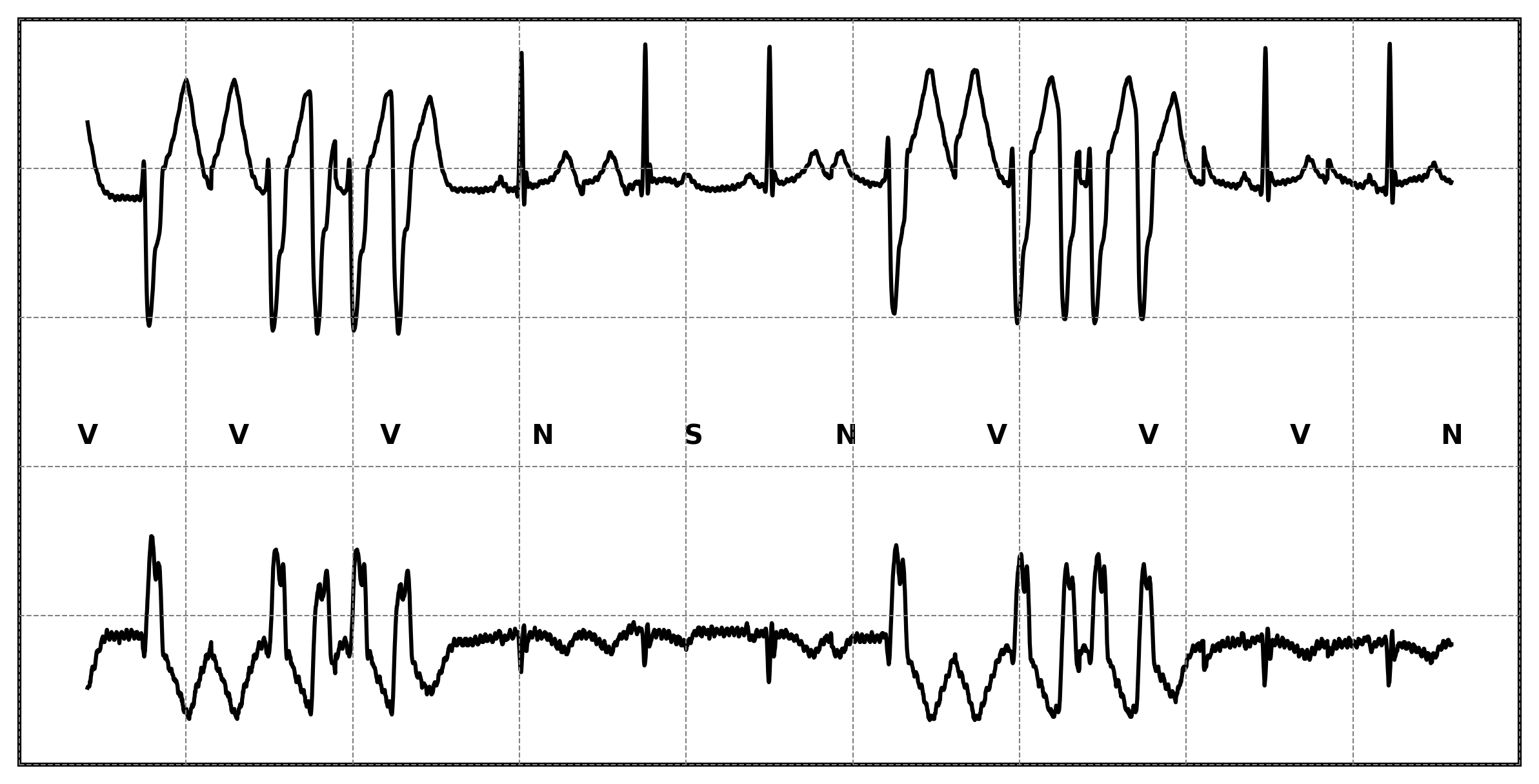}
            \caption{Annotation example from the MIT-BIH database.}
            \label{fig:mit_annotations}
        \end{figure}

        The AAMI-recommended databases include annotations by beat class and fiducial points, such as the R-point, which marks the maximum amplitude of the heartbeat. These annotations are essential for developing automated methods for arrhythmia classification. Figure \ref{fig:mit_annotations} presents an example of annotations according to the AAMI standard. Figure~\ref{fig:mit_annotations} shows a ten-second excerpt from record 205 of the MIT-BIH Database, with lead II at the top, lead V1 at the bottom, and several beat annotations (N, V, A, T) in the center. Each beat type will be described in detail later in this section.
        
        The AAMI standard specifies that only certain types of heartbeats need to be detected by the equipment and methods. It defines 15 beat classes, organized into five major superclasses: Normal (N), Supraventricular ectopic beat (SVEB), Ventricular ectopic beat (VEB), Fusion beat (F), and Unknown beat (Q). The standard also recommends excluding records from patients with pacemakers and segments containing ventricular flutter or fibrillation (VF) from the analysis. Table \ref{tab:beat_types} lists these 15 classes with their symbols and the corresponding superclass hierarchy.
        
        \begin{table}[!htb]
            \centering
            \caption{Main types of beats present in the MIT-BIH database.}
            \label{tab:beat_types}
                \begin{tabular}{|c|c|c|}
                    \hline
                    Group & Symbol & Class \\ \hline
                    \multirow{5}{*}{\begin{tabular}[c]{@{}c@{}}\textbf{N}\\ Normal beat\end{tabular}} & N or . & Normal beat \\ \cline{2-3} 
                     & L & Left bundle branch block beat \\ \cline{2-3} 
                     & R & Right bundle branch block beat \\ \cline{2-3} 
                     & e & Atrial escape beat \\ \cline{2-3} 
                     & j & Nodal (junctional) escape beat \\ \hline
                    \multirow{4}{*}{\begin{tabular}[c]{@{}c@{}}\textbf{SVEB}\\ \\ Supraventricular ectopic beat\end{tabular}} & A & Atrial premature beat \\ \cline{2-3} 
                     & a & Aberrated atrial premature beat \\ \cline{2-3} 
                     & J & Nodal (junctional) premature beat \\ \cline{2-3} 
                     & S & Supraventricular premature beat \\ \hline
                    \multirow{2}{*}{\begin{tabular}[c]{@{}c@{}}\textbf{VEB}\\ Ventricular ectopic beat\end{tabular}} & V & Premature ventricular contraction \\ \cline{2-3} 
                     & E & Ventricular escape beat \\ \hline
                    \begin{tabular}[c]{@{}c@{}}\textbf{F}\\ Fusion beat\end{tabular} & F & Fusion of ventricular and normal beat \\ \hline
                    \multirow{3}{*}{\begin{tabular}[c]{@{}c@{}}\textbf{Q}\\ Unknown beat\end{tabular}} & P or / & Paced beat \\ \cline{2-3} 
                     & f & Fusion of paced and normal beat \\ \cline{2-3} 
                     & U & Unclassifiable beat \\ \hline
                \end{tabular}
        \end{table}
        
        The AAMI also recommends several metrics for evaluating the performance of algorithms in analyzing ECG signals. These metrics are widely used in ECG signal processing, providing standardized benchmarks for comparing different algorithms.
        
        The key metrics recommended by the AAMI include Sensitivity (Se), Positive Predictive Value (+P), False Positive Rate (FPR), and Global Accuracy (Acc). Sensitivity and Positive Predictive Value are also commonly referred to in the literature as Recall and Precision, respectively. Overall accuracy, however, can be heavily biased by the majority class, especially in imbalanced datasets like the MIT-BIH, which contain varying proportions of each beat type. Consequently, Sensitivity, Positive Predictive Value, and False Positive Rate are prioritized as the most relevant metrics for evaluating methods in this context, as they offer a clearer perspective on the classifier's performance across the different beat classes.

    \subsection{MIT-BIH database}
    \label{sec:mit}
    
        The MIT-BIH Arrhythmia Database comprises 48 ECG recordings sampled at 360 Hz, each approximately 30 minutes long. These recordings were collected at the Arrhythmia Laboratory of Beth Israel Hospital between 1975 and 1979 \cite{moody_impact_2001}. The dataset includes recordings from 47 unique patients, with two recordings taken from each.
        
        The dataset represents a diverse group of 25 men, aged between 32 and 89, and 22 women, aged between 23 and 89, with around 60\% of the subjects being hospital inpatients. Each recording includes signals from two ECG leads. Due to anatomical differences and surgical dressings, consistent electrode positioning was not feasible across all subjects. In most cases, one channel records a modified lead II (chest electrode placement), commonly used in ambulatory ECGs, while the second channel typically captures a modified V1 lead. Depending on the individual, however, this secondary lead may vary and could be V2, V4, or V5 \cite{goldberger2000physiobank}. Generally, lead II is preferred for beat detection because of its prominent QRS complex, whereas the secondary lead helps classify specific arrhythmias, such as supraventricular ectopic beats (SVEB) and ventricular ectopic beats (VEB) \cite{goldberger2000physiobank}. The database also includes recordings that capture rare conditions, adding complexity to classification tasks and broadening the variety of cases within the dataset.
        
        Two experienced cardiologists reviewed each recording in the MIT-BIH database independently. They enhanced beat labels, corrected any false detections, reclassified abnormal beats where needed, and included labels for rhythm and signal quality. Following this initial review, the cardiologists compared their annotations and resolved any discrepancies to ensure consensus on each heartbeat's classification. The final annotations were created in accordance with AAMI guidelines.
        
        The beat types present in the MIT-BIH Arrhythmia Database encompass all classes defined by the AAMI (see Table \ref{tab:beat_types}). Table \ref{tab:beat_class_size} shows the distribution of beats across the 48 recordings in the MIT-BIH database.
        
        \begin{table}[!ht]
            \centering
            \caption{Splitting Beat Types in the MIT-BIH Database.}
            \label{tab:beat_class_size}
            \begin{tabular}{|c|c|c|}
                \hline
                AAMI Group & \% of total & Number of beats \\ \hline
                Normal & 82.78 & 90,631 \\ \hline
                SVEB & 2.54 & 2,781 \\ \hline
                VEB & 6.60 & 7,236 \\ \hline
                F & 0.73 & 803 \\ \hline
                P (pacemaker) & 7.32 & 8,010 \\ \hline
                Q & 0.03 & 33 \\ \hline
                Total & 100 & 109,494 \\ \hline
            \end{tabular}
        \end{table}

\section{Inter-patient Paradigm for Data Partitioning}
\label{sec:chazal}

    The inter-patient paradigm proposed by De Chazal \textit{et al.}  \cite{de_chazal_automatic_2004} is a critical approach to data partitioning, designed to ensure generalization and minimize bias in heartbeat classification models. This paradigm requires that heartbeats from individual patients be included exclusively in either the training or testing datasets, preventing overlap that could lead to overfitting to patient-specific patterns. Such overlap can result in artificially inflated model performance, as classifiers may learn unique features specific to individual patients rather than generalizable characteristics of arrhythmias. 
    
    The MIT-BIH arrhythmia database, which includes a variety of arrhythmias and normal sinus rhythms, was selected as the study’s primary data source. To adhere to ANSI/AAMI standards, recordings containing paced beats (102, 104, 107, and 217) should be excluded from the dataset. Paced beats, influenced by artificial pacemakers, exhibit unique morphological and rhythmic characteristics that are not representative of natural cardiac activity. Including these beats could bias the classifier, as they do not align with the typical patterns of naturally occurring arrhythmias.
    
    By strictly separating patient data, the inter-patient paradigm forces models to focus on universally applicable features, significantly improving their ability to classify heartbeats from unseen patients accurately and ensuring clinically relevant and reliable performance.
    
    The AAMI standard provides protocols for testing and evaluating arrhythmia classification methods, including recommendations for databases. However, it does not specify how to partition patients or heartbeats for training and testing, which can result in biased evaluations. De Chazal \textit{et al.}  \cite{de_chazal_automatic_2004} demonstrated that including heartbeats from the same patient in both phases introduces bias, as models memorize patient-specific characteristics during training, leading to inflated test performance (often approaching 100\%). This approach, referred to in the literature as the intra-patient scheme or paradigm, does not reflect real-world clinical conditions where automated arrhythmia detection methods must handle heartbeats from entirely new patients.
    
    To address this limitation, De Chazal \textit{et al.}  \cite{de_chazal_automatic_2004} proposed the inter-patient paradigm, a more realistic data partitioning method. They divided the MIT-BIH database into two datasets: DS1, which includes all heartbeats from records 101, 106, 108, 109, 112, 114, 115, 116, 118, 119, 122, 124, 201, 203, 205, 207, 208, 209, 215, 220, 223, and 230, and DS2, which consists of heartbeats from records 100, 103, 105, 111, 113, 117, 121, 123, 200, 202, 210, 212, 213, 214, 219, 221, 222, 228, 231, 232, 233, and 234.
    
    In this division, DS1 was used exclusively to train the classification model, while DS2 was reserved for evaluation. This ensured that the model had no prior exposure to heartbeats from DS2, as the two datasets comprised recordings from entirely different individuals. The inter-patient paradigm, as implemented by De Chazal \textit{et al.}  \cite{de_chazal_automatic_2004}, better reflects clinical scenarios. The MIT-BIH database was selected for this partitioning because it is the only database recommended by AAMI that includes all five arrhythmia superclasses.
    
    According to De Chazal \textit{et al.}  \cite{de_chazal_automatic_2004}, the initial division of records separated odd and even-numbered entries. To balance class distribution, some records were exchanged between the datasets. As shown in Table \ref{tab:heartbeat_distribution}, both datasets maintain a similar distribution of heartbeats across classes, with each containing approximately 100,000 heartbeats in total. Notably, records 201 and 202, originating from the same patient, are placed in different datasets, while all other records pertain to unique patients.
    
    \begin{table}[!htb]
        \centering
        \caption{Distribution of heartbeats across classes in the datasets as outlined by de Chazal \textit{et al.}  \cite{de_chazal_automatic_2004}.}
        \label{tab:heartbeat_distribution}
        \begin{tabular}{|c|c|c|c|c|c|c|}
            \hline
            Set       & N      & SVEB  & VEB   & F   & Q  & Total   \\ \hline
            DS1       & 45,866 & 944   & 3,788 & 415 & 8  & 51,021  \\ \hline
            DS2       & 44,259 & 1,837 & 3,221 & 388 & 7  & 49,712  \\ \hline
            DS1 + DS2 & 90,125 & 2,781 & 7,009 & 803 & 15 & 100,733 \\ \hline
        \end{tabular}
    \end{table}
    
    De Chazal \textit{et al.}  \cite{de_chazal_automatic_2004} concluded that using DS1 for training and DS2 for testing provides a more realistic evaluation of classification models. This partitioning method increases the difficulty of the task, often reducing classifier performance. Additionally, minority classes such as supraventricular ectopic beats (SVEB) and ventricular ectopic beats (VEB)—which represent some of the most clinically critical arrhythmias—are disproportionately affected by this protocol, posing significant challenges for robust model evaluation.

    Beyond the specific paradigm proposed by De Chazal, an approach is also considered inter-patient if there is no overlap of heartbeats from the same patient between training and testing. This ensures that the neural network is evaluated only on entirely unseen data, providing a stricter assessment of its generalization ability. Thus, datasets other than the MIT-BIH Arrhythmia Database can be used, as long as heartbeats from the same patient do not appear in both training and testing phase.

\section{Implementing Neural Networks on Low-Cost Hardware for ECG Classification: Challenges and Optimization Techniques}
\label{sec:hardware}

    Deploying deep learning models on low-power hardware, such as microcontrollers and edge devices, presents significant challenges due to their limited computational resources, memory constraints, and stringent energy efficiency requirements. These devices are integral to applications like wearable health monitors, IoT sensors, and portable medical equipment. They provide real-time monitoring of heart activity and allow early detection of arrhythmias or other cardiac events without requiring hospitalization or large stationary equipment. Additionally, they enable monitoring in resource-limited settings, making quality healthcare diagnostics more accessible in rural or underserved areas.  
    
    As portable and wearable medical monitoring applications grow, ongoing research is focused on improving neural network efficiency and accuracy for low-cost hardware. The primary challenge is the substantial computational demand for deep learning models, which often exceeds the capabilities of low-power hardware. Additionally, the memory footprint of these models can surpass the available memory on such devices, leading to issues such as slow inference times and increased power consumption. 
    
    To address these challenges, several model optimization techniques are employed to adapt neural networks for deployment on low-cost hardware, such as Pruning \cite{rawal2023hardware,le2023lightx3ecg,ran2022homecare,lu2022efficient}, Quantization \cite{ran2022homecare,ribeiro2022ecg}, lightweight network architectures \cite{saadatnejad2019lstm,meng2022enhancing,alamatsaz2024lightweight,chen2024novel,le2023lightx3ecg,ribeiro2022ecg,tesfai2022lightweight,xing2022accurate,sakib2021proof,jeon2020lightweight,feyisa2022lightweight}, FPGA acceleration \cite{rawal2023hardware,chu2022neuromorphic,ran2022homecare,xing2022accurate,tsoutsouras2017exploration,lu2022efficient,zhang2023configurable,cheng2022efficient,lu2021efficient}, and On-chip Learning \cite{mao2022ultra}. 

    In the following subsections, we will discuss the applicability of each technique to ECG classification, highlighting studies from our review that implemented these methods and analyzing their performance.
    
    \subsection{Pruning Techniques}
        Pruning is employed to minimize the model footprint by deactivating neurons that contribute minimally to the classification task, effectively streamlining the model without compromising its performance \cite{blalock2020state}. 

        Among the analyzed articles, some employed pruning techniques. One such study, Le \textit{et al.} \cite{le2023lightx3ecg}, presents an approach to ECG classification using only three leads (I, II, and V1), tailored for deployment in resource-constrained environments. The model features redesigned 1D-SEResNet backbones, a Lead-wise Attention module for effective feature aggregation, and Lead-wise Grad-CAM for enhanced interpretability, achieving both high accuracy and clinical relevance. To optimize the model for portable and wearable devices, pruning was applied post-training, removing 80\% of low-impact weights based on $L_1$-norm. This reduced memory usage by threefold without compromising performance, making the model highly efficient for low-power hardware. Le \textit{et al.} \cite{le2023lightx3ecg} achieved an F1-score of 0.9718 and accuracy of 98.73\% on the Chapman dataset, and an F1-score of 0.8004 with 96.28\% accuracy on the CPSC-2018 dataset. Compared to state-of-the-art methods, it outperformed in classification metrics while reducing computational costs (1.34B FLOPs) and storage requirements (6.52 MB after pruning). The use of Lead-wise Grad-CAM further enhanced the model’s interpretability, aligning its predictions with clinical diagnostic standards.

        Rawal \textit{et al.} \cite{rawal2023hardware} present a hardware-based approach for ECG arrhythmia detection using a 1D-CNN architecture optimized for low-power devices. The study focuses on developing a safety-critical system for arrhythmia classification, specifically targeting Atrial Fibrillation (AF), by employing pruning and other hardware-efficient techniques. Pruning was applied to reduce computational complexity and enhance hardware efficiency. Starting with a highly accurate Software-Selected CNN Architecture (SSCA), the authors used $L_1$-norm pruning to remove filters with minimal contribution to classification accuracy. Each pruning iteration was followed by retraining for 50 epochs to recover any performance losses, resulting in a Pruned CNN Architecture (PRCA) with fewer filters and hardware-ready efficiency.

        The PRCA achieved 90.80\% overall accuracy in software implementation for the 2017 PhysioNet/Computing in Cardiology Challenge dataset \cite{clifford2017af}, with 98.70\% accuracy specifically for AF classification. On hardware, implemented on a ZYNQ Ultrascale FPGA, the model demonstrated an accuracy of 86.37\%, a slight decrease attributed to fixed-point quantization and approximation techniques. The pruning process reduced computational complexity by 48.92\% compared to the Supreme CNN Architecture (SCA), which was the most accurate but computationally intensive model in the study. Additionally, the PRCA on FPGA consumed only 628 mW of power.

        Lu \textit{et al.} \cite{lu2022efficient} focus on developing an efficient CNN accelerator for wearable ECG classification using unstructured pruning. This technique reduces unnecessary connections in the neural network based on their contribution, achieving a high sparsity level of 70\% while maintaining a classification accuracy of 98.99\% in the PTB-XL dataset \cite{wagner2020ptb}. To optimize the performance of pruned models, the researchers proposed a novel tile-first dataflow and a compressed data storage format. These innovations improve computational efficiency by skipping zero-weight multiplications, effectively utilizing limited hardware resources in wearable devices.

        The hardware implementation of the pruned 1D-CNN model demonstrated minimal accuracy loss (0.1\%) when mapped onto an FPGA platform. It achieved a computing efficiency of 118.75\% and consumed just 3.93 $\mu\text{J}$ per classification, outperforming dense models in both efficiency and energy consumption. Additionally, the model's hardware feasibility was validated for real-time ECG beat classification, making it highly suitable for resource-constrained wearable devices.

    \subsection{Quantization Techniques}
    
        Quantization is a technique that reduces the bit-width of model parameters, often transitioning from 32-bit floating-point representations to lower bit-width formats like 8-bit integers \cite{nagel2021white}. This approach significantly decreases memory usage and accelerates computations, as lower-bit operations require less processing power and energy \cite{nagel2021white}. 

        Some of the analyzed articles use Quantization techniques, such as the study by Ran \textit{et al.} \cite{ran2022homecare}, which presented an innovative ECG diagnostic model for real-time, multilabel classification of 12-lead ECG signals. Using a dataset of 206,468 recordings to classify 26 rhythm types, the model integrates a deep convolutional neural network (DCNN) with hardware co-optimization techniques, achieving significant compression and performance improvements for embedded platforms.

        To optimize the model for resource-constrained environments, the study employs channel-level pruning and integer quantization. Pruning compresses the model by converting 32-bit floating-point parameters to 8-bit integers without compromising accuracy. 
        
        The optimized model was deployed on an FPGA-based platform, achieving a diagnostic latency of 2.895 seconds per sample with a power consumption of only 2.81 W. Performance evaluations demonstrated an average F1 score of 0.913 and exact match ratios of 86.7\% and 80\% on internal and external datasets, respectively, surpassing cardiologist-level benchmarks.

        Ribeiro \textit{et al.} \cite{ribeiro2022ecg} proposed an energy-efficient and accurate approach to classifying arrhythmias using raw ECG signals. The model is a lightweight 1D convolutional neural network optimized for deployment on low-power edge devices. The authors used post-training 8-bit integer quantization to compress the network, reducing memory and computational requirements.

        The model was trained using the MIT-BIH Arrhythmia Database, incorporating five AAMI-recommended heartbeat classes. To handle data imbalance, random oversampling was employed, creating a balanced dataset for training and validation. The quantized model achieved a classification accuracy of 99.6\%, with sensitivity and specificity consistently exceeding 99\% across all classes. However, they do not follow the guidelines specified by De Chazal \textit{et al.} \cite{de_chazal_automatic_2004}, which artificially inflates the classification metrics.
        
        For deployment feasibility, the quantized model demonstrated significant reductions in size and inference time, achieving an average inference time of 7.65 ms and energy consumption of 5.85 mJ per prediction on an ARM Cortex A55 processor. The model's total size was reduced to 93 kB.

    \subsection{Lightweight Network Architectures}
    
        Lightweight Network Architectures, such as MobileNet \cite{sinha2019thin}, SqueezeNet \cite{koonce2021squeezenet}, and TinyML \cite{lin2022ondevice}, are specifically crafted for constrained devices. These architectures integrate techniques such as pruning, quantization, and parameter reduction to optimize their performance while maintaining efficiency on resource-limited hardware.

        Within the reviewed articles, the study by Saadatnejad \textit{et al.} \cite{saadatnejad2019lstm} proposed a lightweight, efficient model for real-time ECG classification on wearable devices with limited computational resources. The model integrates wavelet transform and two LSTM-based recurrent neural network models ($\alpha$ and $\beta$), designed to process ECG signals along with extracted RR interval and wavelet features. By using two smaller RNN models instead of a single large model, the system reduces computational demands.

        The model achieved patient-specific training by combining global ECG data, containing representative arrhythmias, with five minutes of patient-specific ECG data, adhering to AAMI standards. Evaluations demonstrate the model's superior performance, particularly in detecting ventricular ectopic beats (VEB) and supraventricular ectopic beats (SVEB), with F1 scores surpassing previous methods by up to 15.5\%.

        Meng \textit{et al.} \cite{meng2022enhancing} proposed the Lightweight Fussing Transformer (LFT) model, an approach designed to address the challenges of processing noisy and dynamic ECG signals from wearable devices. The LFT model replaces the traditional self-attention mechanism of transformers with a more efficient Lightweight Convolution Attention (LCA), which utilizes depth-wise convolutions to reduce computational complexity and parameter size while maintaining robust attention to critical ECG features. The architecture also includes a CNN-based input structure with a local channel attention mechanism to enhance the extraction of morphological features from heartbeats.

        The model integrates a two-level attention mechanism, combining local attention to focus on intra-heartbeat features and global attention to capture inter-beat dependencies. Additionally, noise reduction techniques, such as Butterworth filters and wavelet transforms, are employed to mitigate baseline drift and electromagnetic interference. Advanced resampling methods are used to address data imbalance during training.
        
        Evaluated on a large-scale dynamic ECG dataset, the LFT model achieved an overall accuracy of 99.32\%, with the following performance in specific classes: Class N (Normal) achieved a precision of 0.9975, sensitivity of 0.9984, and F1-score of 0.9979; Class S (Supraventricular Premature Beat) recorded a precision of 0.9386, sensitivity of 0.8300, and F1-score of 0.8810; and Class V (Premature Ventricular Contraction) achieved a precision of 0.9158, sensitivity of 0.9447, and F1-score of 0.9300. The LFT model also demonstrated a parameter reduction of 72\% compared to traditional self-attention architectures, emphasizing its suitability for deployment in wearable devices with limited computational resources.

        Sakib \textit{et al.} \cite{sakib2021proof} introduced DL-LAC, a lightweight arrhythmia classification model specifically designed for resource-constrained IoT sensors. The model utilizes a one-dimensional convolutional neural network (1D-CNN) to process raw single-lead ECG signals, eliminating the need for noise filtering and manual feature extraction. Compliant with AAMI standards, DL-LAC classifies four heartbeat types: normal, supraventricular ectopic, ventricular ectopic, and fusion beats. 

        The proposed lightweight architecture integrates convolutional layers for automated feature extraction and fully connected layers for classification, employing dropout and max-pooling for regularization and data compression. This design minimizes resource consumption. DL-LAC achieved an accuracy rate of 96.67\% across datasets. Deployment tests on hardware platforms like Raspberry Pi and Jetson Nano confirmed its suitability for ultra-edge computing, exhibiting low execution times and minimal memory usage. 

    \subsection{FPGA acceleration}
        
        FPGA acceleration provides an additional solution by offering reconfigurable chips capable of highly efficient processing \cite{muthuramalingam2008neural}. They allow network-specific architectures to be designed, resulting in significant speed and efficiency gains compared to general-purpose processors \cite{muthuramalingam2008neural}. FPGA-based accelerators support parallel operations that align well with convolutional neural networks (CNNs), which are often used in ECG classification. Although FPGAs are more costly than standard microcontrollers, they remain affordable compared to GPUs and are valuable in applications where real-time processing and minimal power consumption are necessary.

        Among the studies analyzed in this work, Rawal \textit{et al.} \cite{rawal2023hardware} presented a hardware-oriented implementation of a 1D-CNN architecture for ECG arrhythmia classification, focusing on atrial fibrillation (AF). The proposed framework addresses the trade-off between accuracy and computational complexity, making it suitable for safety-critical systems and resource-constrained environments.

        The study developed three CNN architectures: Supreme CNN Architecture (SCA), Software-Selected CNN Architecture (SSCA), and Pruned CNN Architecture (PRCA). Among these, the PRCA was optimized for hardware implementation using pruning techniques, reducing computational complexity by 48.92\% while maintaining an overall classification accuracy of 90.8\%. For AF classification, the accuracy was 97.34\%. The system was implemented on a ZYNQ Ultrascale FPGA, achieving a low power consumption of 628 mW, which is 66.9\% less than comparable methods.

        Xing \textit{et al.} \cite{xing2022accurate} presented a lightweight, energy-efficient model for ECG classification, optimized for real-time use on portable devices. The research leverages a Spiking Neural Network (SNN) enhanced by a Channel-Wise Attentional Mechanism (CAM) to classify five heartbeat categories (N, S, V, F, and Q) from the MIT-BIH Arrhythmia Database.

        The FPGA implementation of the ECG classification model leverages the event-driven nature of Spiking Neural Networks (SNNs) with Leaky Integrate-and-Fire (LIF) neurons, enabling computations only when spikes occur, significantly reducing power consumption. Utilizing FPGA parallelism, the model incorporates a Channel-Wise Attentional Mechanism (CAM) for dynamic feature weighting and employs fixed-point arithmetic and pipelined architecture for efficient processing. Optimized memory management ensures minimal reliance on external memory, enhancing speed and energy efficiency. The system achieves energy consumption of 346.33 $\mu\text{J}$ per beat and a runtime of 1.37 ms per classification, making it highly suitable for real-time ECG monitoring in wearable and portable devices.
        
        The experimental results show an overall accuracy of 98.26\%, a sensitivity of 94.75\%, and an F1-score of 89.09\% on the MIT-BIH database and under 70:30 partition. For inter-patient evaluation, the model achieved an accuracy of 92.07\%. The energy efficiency is notable, with an energy consumption of 346.33 $\mu\text{J}$ per beat and a runtime of 1.37 ms, making it suitable for embedded platforms.

        Zhang \textit{et al.} \cite{zhang2023configurable} proposed a 1-D convolutional neural network (CNN)-based inference engine tailored for ECG classification in mobile healthcare devices. The proposed system incorporates a highly configurable architecture, leveraging systolic arrays and multi-level data buffers to minimize power consumption and enhance flexibility. It supports multiple convolutional and fully connected layer configurations, enabling efficient classification across diverse ECG scenarios.

        For hardware evaluation, the design was implemented on an FPGA (Xilinx Zynq-7000 SoC ZC706 evaluation board) operating at 200 MHz. The system was tested with two 1-D CNN models: a lightweight energy-efficient model and a high-accuracy model, both trained on the MIT-BIH arrhythmia database. The lightweight model demonstrated a classification accuracy of 98.9\% with an energy consumption of 2.05 $\mu\text{J}$ per inference, while the high-accuracy model achieved a 99.3\% classification accuracy at 14.27 $\mu\text{J}$ per inference.

    \subsection{On-device Learning}

        On-chip learning enables real-time adaptation of models directly on embedded hardware \cite{dhar2021survey}. Unlike traditional approaches where models are trained in cloud-based infrastructures or on powerful computing servers and deployed as static inference engines, on-chip learning allows models to be fine-tuned locally \cite{dhar2021survey}. This capability is particularly advantageous in scenarios where continuous adaptation to new data is necessary, such as wearable and implantable medical devices \cite{tekin2024review}.

        A key challenge of on-chip learning is the severe constraint on computational resources and memory in embedded systems. Training deep learning models requires additional memory for storing gradients and intermediate activations, which significantly exceeds the capacity of most microcontrollers. Recent advancements, such as those presented by Lin \textit{et al.} \cite{lin2022ondevice}, proposed an algorithm-system co-design framework that enables on-device training with as little as 256KB of memory. Their method introduces Quantization-Aware Scaling (QAS), which automatically calibrates the gradient scales to stabilize training for 8-bit quantized models, effectively matching the accuracy of floating-point training without extra memory overhead. Additionally, they employ Sparse Update techniques, which reduce memory usage by selectively skipping gradient computations for less important layers and sub-tensors. These innovations are implemented within a lightweight training system called the Tiny Training Engine (TTE), which offloads the backpropagation computation to compile-time, significantly reducing runtime overhead. This framework enables tiny on-device training of convolutional neural networks under the strict constraints of 256KB SRAM and 1MB Flash, achieving up to a 1000× reduction in memory usage compared to traditional deep learning frameworks like PyTorch and TensorFlow, while maintaining high accuracy in applications like Visual Wake Words (VWW).
        
        An approach for ECG classification is proposed by Mao et al. \cite{mao2022ultra}, who addressed this issue by proposing an ultra-energy-efficient ECG classification processor that incorporates Spiking Neural Networks (SNNs) for real-time adaptation. Their approach leverages event-driven processing, reducing power consumption while maintaining classification accuracy. The processor designed by the authors integrates a spike-based feature extraction mechanism that dynamically adapts to variations in ECG signals, allowing real-time adjustments without requiring full backpropagation-based retraining. By leveraging a neuromorphic computing paradigm, the model updates only relevant parameters through local weight adjustments, significantly reducing the computational overhead compared to traditional gradient-based methods. Additionally, their implementation optimizes memory usage by employing sparsity-aware architectures that minimize redundant computations.

\section{Results of the Systematic Literature Review}
\label{sec:resultados}

    This section presents the main findings of our systematic review, focusing on key aspects of arrhythmia classification models and their alignment with established standards and practices. The analysis covers multiple dimensions, including compliance with the AAMI recommendations, data partitioning methodologies, performance metrics, and the suitability of models for deployment on low power hardware.

    \subsection{Adherence of Selected Studies to E3C}
\begin{figure}[!htb]
            \centering
            \includegraphics[width=0.6\linewidth]{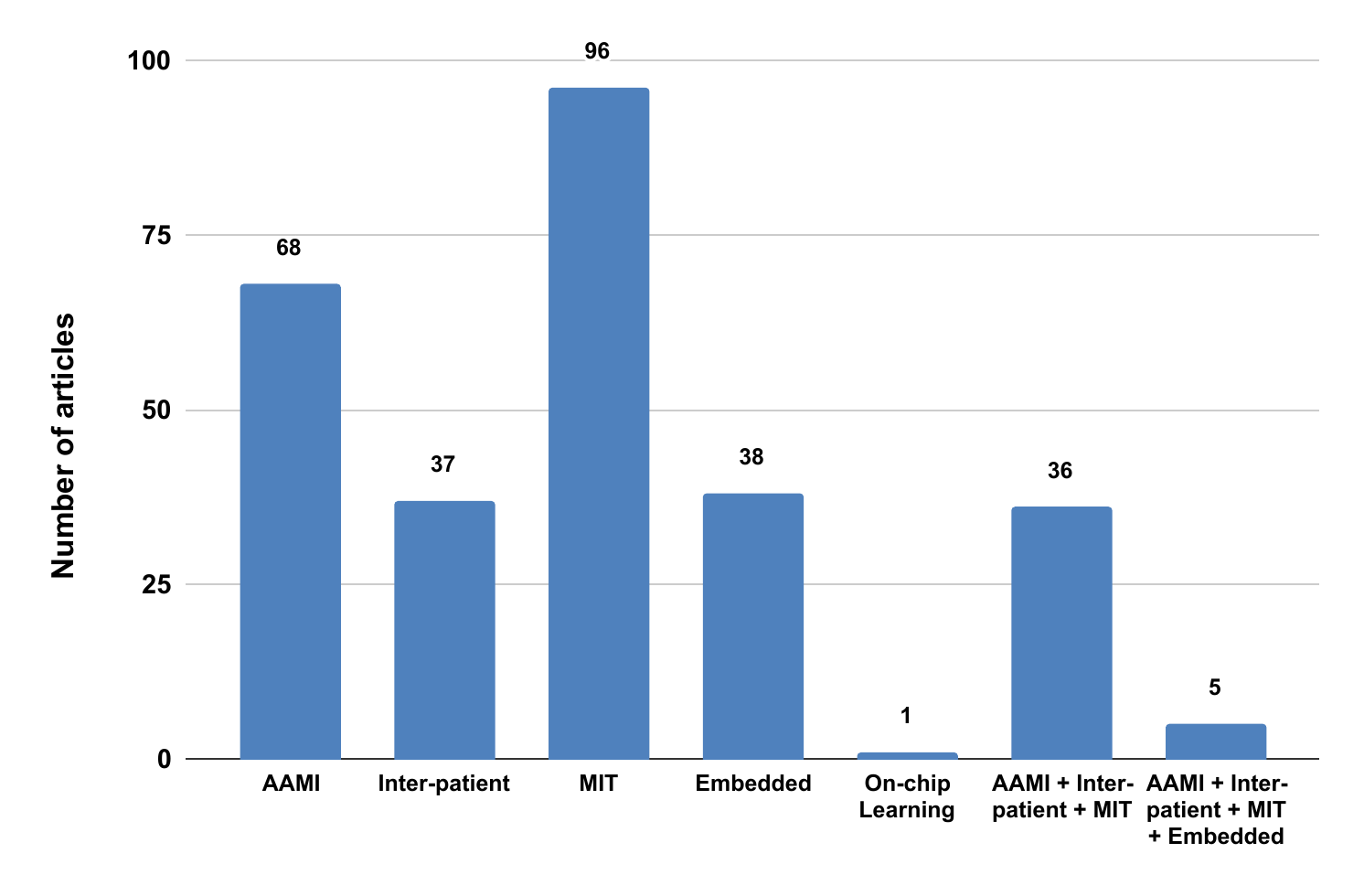}
            \caption{Distribution of the 122 selected articles based on adherence to AAMI standards, use of the inter-patient paradigm by De Chazal\textit{ et al.} \cite{de_chazal_automatic_2004}, utilization of the MIT-BIH database, and consideration of hardware feasibility. The bars represent the number of articles meeting each criterion, along with their respective proportions in the dataset.}
            \label{fig:criteria_analysis}
        \end{figure}

        Figure \ref{fig:criteria_analysis} provides a detailed distribution of the 122 selected articles, highlighting their adherence to key criteria in arrhythmia classification research. Among these, 68 articles (55.7\%) complied with AAMI guidelines, 37 articles (30.3\%) adhered to the inter-patient paradigm, 96 articles (78.7\%) utilized the MIT-BIH Arrhythmia Database, 38 (31.1\%) articles considered embedded feasibility, and only 5 (4.1 \%) articles adhered to E3C. 
        
        In order to further analyze these articles, we developed Tables \ref{tab:table1}, \ref{tab:table2}, \ref{tab:table3}, \ref{tab:table4}, and \ref{tab:table5}, which provide a comprehensive summary of the 122 articles selected after the initial evaluation stages discussed in Section \ref{sec:methodology}. The ``Article'' field lists the reference to the analyzed study. The ``Classes'' field specifies the number of heartbeat classes considered, offering insight into the classification complexity. The ``Follow AAMI'' field indicates whether the study adheres to the standards established by the AAMI. The ``Inter-patient'' field indicates whether the study implements the inter-patient paradigm proposed by De Chazal \textit{et al.} \cite{de_chazal_automatic_2004} or an alternative inter-patient approach., an essential methodology for robust and unbiased model evaluation. The ``MIT-BIH'' field indicates whether the study utilized the MIT-BIH Arrhythmia Database, the most commonly used dataset among those recommended by AAMI and a standard benchmark in the ECG classification field. The ``Classifier'' field describes the type of machine learning model or algorithm employed. The ``Embedded'' field indicates whether the study evaluated the feasibility of deploying the model on low-power hardware. Finally, the ``Performance'' field summarizes key metrics such as accuracy, sensitivity, specificity, and F1-score.

        \begin{table}[!htp]
            \centering
            \caption{Comparative summary of studies on arrhythmia classification: characteristics of analyzed articles, including adherence to AAMI recommendations, use of the inter-patient paradigm, database used, classifier model, embedded system feasibility, on-chip learning capability, and reported performance metrics (Part 1 of 5).}
            \label{tab:table1}
            \scriptsize
            \renewcommand{\arraystretch}{1.1}
            \begin{tabularx}{\linewidth}{|>{\centering\arraybackslash}p{0.6cm}|>{\centering\arraybackslash}p{0.8cm}|>{\centering\arraybackslash}p{0.8cm}|>{\centering\arraybackslash}p{0.8cm}|>{\centering\arraybackslash}p{0.8cm}|>{\centering\arraybackslash}p{2cm}|>{\centering\arraybackslash}p{1cm}|>{\centering\arraybackslash}p{1cm}|X|}
                \hline
                \rowcolor[HTML]{CCCCCC}
                \textbf{Article} & \textbf{Classes} & \textbf{Follow AAMI} & \textbf{Inter-patient} & \textbf{MIT-BIH} & \centering \textbf{Classifier} & \textbf{Embedded} & \textbf{On-Chip Learning} & \textbf{Performance} \\ 
                \hline
                \rowcolor[HTML]{F2F2F2} \cite{hannun2019cardiologist} &12 &No &No &No &DNN with 34 layers &No &No &AUC: 0,91 for all classes \newline mean F1-score: 0,837 \\ 
                \hline
                \rowcolor[HTML]{FFFFFF} \cite{yildirim2018arrhythmia} &13, 15 and 17 &No &No &Yes &1D-CNN with 16 layers &No &No &13 Classes: SEN = 93,52\%, SPE = 99,61\%, F1-Score = 92,45\%, ACC = 95,2\% 15 Classes: SEN = 88,57\%, SPE = 99,39\%, F1-Score = 89,28\%, ACC = 92,51\% 17 Classes: SEN = 83,91\%, SPE = 99,41\%, F1-Score = 85,38\%, ACC = 91,33\% \\ 
                \hline
                \rowcolor[HTML]{F2F2F2} \cite{oh2018automated} &5 &No &No &Yes &CNN with LSTM &No &No &ACC: 98,10\% Sen: 97,50\% Spe: 98,70\% \\ 
                \hline
                \rowcolor[HTML]{FFFFFF} \cite{yao2020multi} &9 &Yes &No &No &Attention-based Time-Incremental Convolutional Neural Network (ATI-CNN) &Yes &No &Precision: 82,6\% Mean Recall: 80,1\% Mean F1-score: 81,2\% \\ 
                \hline
                \rowcolor[HTML]{F2F2F2} \cite{sannino2018deep} &5 &Yes &Yes &Yes &DNN with 7 layers &No &No &Acc: 99,09\% Sen: 98,55\% Spe: 99,52\% \\ 
                \hline
                \rowcolor[HTML]{FFFFFF} \cite{saadatnejad2019lstm} &5 &Yes &No &Yes &LSTM, RNN, Wavelet transform and MLP &Yes &No &SVEB: Acc: 93,78\% (360 Hz) and 93,63\% (114 Hz) Sen: 88,39\% (360 Hz) and 88,62\% (114 Hz) PPV: 33,63\% (360 Hz) and 35,49\% (114 Hz) FPR: 6,68\% (360 Hz) and 6,17\% (114 Hz) VEB : Acc: 96,94\% (360 Hz) and 95,87\% (114 Hz) Sen: 85,22\% (360 Hz) and 85,54\% (114 Hz) PPV: 56,63\% (360 Hz) and 60,83\% (114 Hz) FPR: 4,11\% (360 Hz) and 3,47\% (114 Hz) \\ 
                \hline
                \rowcolor[HTML]{F2F2F2} \cite{yildirim2019new} &5 &No &No &Yes &Deep Convolutional Autoencoder and LSTM &No &No &Mean Acc: 99,23\% Precision: 0,99 Recall: 0,99 F1-score: 0,99 \\ 
                \hline
                \rowcolor[HTML]{FFFFFF} \cite{mathews2018novel} &5 &Yes &Yes &Yes &Restricted Boltzmann Machine (RBM) and Deep Belief Networks (DBN). &Yes &No &SVEB: Acc: 93,63\% (114 Hz) Sen: 88,62\% PPV: 35,49\% FPR: 6,17\% VEB: Acc: 95,87\% (114 Hz) Sen: 85,54\% PPV: 60,83\% FPR: 3,47\% \\ 
                \hline
                \rowcolor[HTML]{F2F2F2} \cite{sahoo2017multiresolution} &4 &No &No &Yes &Neural Network and SVM &No &No &With SVM: Acc: 98,39\% Sen: 96,86\% Spe: 98,92\% With ANN: Acc: 96,67\% Sen: 93,57\% Spec: 97,75\% \\ 
                \hline
                \rowcolor[HTML]{FFFFFF} \cite{tuncer2019automated} &17 &Yes &Yes &Yes &1-Nearest Neighbor (1NN) and Discrete Wavelet Transform (DWT) &No &No &Acc: 95,0\% \\ 
                \hline
                \rowcolor[HTML]{F2F2F2} \cite{xu2018towards} &5 &Yes &Yes &Yes &DNN &No &No &Acc: 94,7\% Sen of S: 77,3\% Sen of V: 93,7\% Spe: 86,7\% \\ 
                \hline
                \rowcolor[HTML]{FFFFFF} \cite{petmezas2021automated} &4 &No &No &Yes &CNN and LSTM &No &No &Sen: 97,87\% Spe: 99,29\% \\ 
                \hline
                \rowcolor[HTML]{F2F2F2} \cite{murat2020application} &5 &Yes &No &Yes &CNN-LSTM &No &No &Acc: 99,11\% \\ \hline
                \rowcolor[HTML]{FFFFFF} \cite{zhai2018automated} &5 &Yes &No &Yes &CNN &No &No &Acc for SVEB: >95\% Sen for S beats: 85,3\% PPR for S beats: 70\% Sen and PPR for V beats: both >90\%. \\ 
                \hline
                \rowcolor[HTML]{F2F2F2} \cite{sellami2019robust} &5 &Yes &Yes &Yes &CNN &No &No &Intra-patient paradigm: Acc: 99,48\%, Sen: 96,97\%, Spec: 99,87\%, PPR: 98,83\%. Inter-patient paradigm: Acc: 88,34\%, Sen: 90,90\%, Spec: 88,51\%, PPR: 48,25\% \\ 
                \hline
                \rowcolor[HTML]{FFFFFF} \cite{hammad2020multitier} &5 &Yes &Yes &Yes &DNN, ResNet-LSTM and Genetic Algorithm (GA). &No &No &Acc: 98,0\%. Sen: 99,7\%. Spec: 98,9\%. PPR: 95,8\%. \\ 
                \hline
                \rowcolor[HTML]{F2F2F2} \cite{oh2019automated} &5 &No &No &Yes &Modified U-Net &No &No &Acc: 97,32\% \\ 
                \hline
                \rowcolor[HTML]{FFFFFF} \cite{yang2018automatic} &5 &Yes &No &Yes &Principal Component Analysis Network (PCANet) and SVM &No &No &Acc: 97,77\% (without noise) e 97,08\% (with noise) \\ 
                \hline
                \rowcolor[HTML]{F2F2F2} \cite{shaker2020generalization} &15 &Yes &No &Yes &CNN and Generative Adversarial Networks (GANs) &No &No &Acc: 98,0\% Prec: 90,0\% Sen: 97,7\% Spe: 97,4\% \\ 
                \hline
                \rowcolor[HTML]{FFFFFF} \cite{alfaras2019fast} &5 &Yes &Yes &Yes &Echo State Networks (ESN) &No &No &MIT-BIH (Lead II): Se = 92,7\%, PPV = 86,1\%. MIT-BIH (Lead V1): Se = 95,7\%, PPV = 75,1\% \\ 
                \hline
                \rowcolor[HTML]{F2F2F2} \cite{mathunjwa2021ecg} &6 &No &No &Yes &CNN &No &No &Acc: 98.41\% Prec: 98.75\% Sen: 97.81\% \\ 
                \hline
                \rowcolor[HTML]{FFFFFF} \cite{wang2021automatic} &5 &Yes &Yes &Yes &Continuous Wavelet Transform (CWT) and CNN &No &No &PPV: 70,75\% Sen: 67,47\% F1-Score: 68,76\% Acc: 98,74\% \\ 
                \hline
                \rowcolor[HTML]{F2F2F2} \cite{zhang2020ecg} &9 &No &No &No &Spatio-Temporal Attention-based Convolutional Recurrent Neural Network (STA-CRNN). &No &No &F1-Score: 83.50\% \\ 
                \hline
            \end{tabularx}
        \end{table}
        
        \begin{table}[!htp]
            \centering
            \caption{Comparative summary of studies on arrhythmia classification: characteristics of analyzed articles, including adherence to AAMI recommendations, use of the inter-patient paradigm, database used, classifier model, embedded system feasibility, on-chip learning capability, and reported performance metrics (Part 2 of 5).}
            \label{tab:table2}
            \scriptsize
            \renewcommand{\arraystretch}{1.1}
            \begin{tabularx}{\linewidth}{|>{\centering\arraybackslash}p{0.6cm}|>{\centering\arraybackslash}p{0.8cm}|>{\centering\arraybackslash}p{0.8cm}|>{\centering\arraybackslash}p{0.8cm}|>{\centering\arraybackslash}p{0.8cm}|>{\centering\arraybackslash}p{2cm}|>{\centering\arraybackslash}p{1cm}|>{\centering\arraybackslash}p{1cm}|X|}
                \hline
                \rowcolor[HTML]{CCCCCC}
                \textbf{Article} & \textbf{Classes} & \textbf{Follow AAMI} & \textbf{Inter-patient} & \textbf{MIT-BIH} & \centering \textbf{Classifier} & \textbf{Embedded} & \textbf{On-Chip Learning} & \textbf{Performance} \\ 
                \hline
                \rowcolor[HTML]{FFFFFF} \cite{xia2018automatic} &4 &Yes &Yes &Yes &Stacked Denoising Autoencoder (SDAE) &No &No &MIT-BIH : VEB: Acc = 99.8\%, Sen = 98.2\%, Spe = 99.9\%, Ppr = 98.4\% SVEB: Acc = 99.8\%, Sen = 94.3\%, Spe = 99.9\%, Ppr = 95.3\% WDDB: VEB: Acc = 99.8\%, Sen = 97.4\%, Spe = 99.9\%, Ppr = 98.3\% SVEB: Acc = 99.8\%, Sen = 93.7\%, Spe = 99.9\%, Ppr = 95.8\% \\ 
                \hline
                \rowcolor[HTML]{F2F2F2} \cite{wang2020deep} &3 and 9 &No &No &No &Multi-Scale Fusion Convolutional Neural Network (DMSFNet) &No &No &CPSC 2018 Dataset: Prec: 83.8\% Recall: 82.2\% mean F1-score: 82.8\% PhysioNet/CinC 2017 Dataset: Prec: 85.6\% Recall: 82.9\% mean F1-score: 84.1\% \\ 
                \hline
                \rowcolor[HTML]{FFFFFF} \cite{weimann2021transfer} &4 &No &No &No &CNN &No &No &Average F1 macro score on the PhysioNet/CinC 2017 dataset after 10 runs: Beat classification: 0.779 ± 0.014 Rhythm classification: 0.767 ± 0.012 Heart rate classification: 0.766 ± 0.011 Future prediction: 0.758 ± 0.013 \\ 
                \hline
                \rowcolor[HTML]{F2F2F2} \cite{niu2019inter} &5 &Yes &Yes &Yes &Multi-Perspective Convolutional Neural Network (MPCNN) &No &No &Acc: 96.4\% F1-score for SVEB: 76.6\% F1-score for VEB: 89.7\% Sen for SVEB: 76.5\% Sen for VEB: 85.7\% \\ 
                \hline
                \rowcolor[HTML]{FFFFFF} \cite{han2020deep} &4 &No &No &No &CNN &No &No &Acc: 88\% F1-Score: 87\% \\ 
                \hline
                \rowcolor[HTML]{F2F2F2} \cite{li2018patient} &5 &Yes &No &Yes &Tuned Dedicated CNN (TDCNN) &Yes &No &Acc: 96.89\% Sen: 95.5\% for VEB and 68.7\% for SVEB. Ppr: 92.2\% for VEB and 94.7\% for SVEB. Spe: 99.1\% for VEB and 99.8\% for SVEB. \\ 
                \hline
                \rowcolor[HTML]{FFFFFF} \cite{romdhane2020electrocardiogram} &5 &Yes &Yes &Yes &CNN &No &No &Ac: 98,41\% F1-Score: 98,38\% Prec: 98,37\% Recall: 98,41\% \\ 
                \hline
                \rowcolor[HTML]{F2F2F2} \cite{al2018convolutional} &5 &Yes &Yes &Yes &CNN and Continuous Wavelet Transform (CWT) &No &No &For VEB: Sen = 99,6\%, Ppr = 99,3\%, Spe = 100\%, Acc = 99,9\%. For SVEB: Sen = 98,3\%, Ppr = 98,9\%, Spe = 100\%, Acc = 99,9\% \\ 
                \hline
                \rowcolor[HTML]{FFFFFF} \cite{kamaleswaran2018robust} &4 &No &No &No &CNN &Yes &No &Acc: 85.99\% F1-Score: Average of 0.83 across all classes. \\ 
                \hline
                \rowcolor[HTML]{F2F2F2} \cite{alarsan2019analysis} &5 &No &No &Yes &Gradient-Boosted Trees (GBT) and Random Forest (RF) &No &No &Acc: 98.03\% Spe: 97\% Sen: 98\% \\ 
                \hline
                \rowcolor[HTML]{FFFFFF} \cite{kiranyaz2017personalized} &5 &Yes &No &Yes &1D CNN &No &No &Acc: 97,75\% Sen: 82,51\% Spe: 99,55\% Ppr: 95,55\% \\ 
                \hline
                \rowcolor[HTML]{F2F2F2} \cite{wang2020high} &5 &Yes &Yes &Yes &Fully Connected Neural Network &No &No &Acc: 93.4\% Sen: 95.1\% (N), 90.3\% (S), 84.1\% (V), 0.25\% (F) Pre: 98.3\% (N), 43.5\% (S), 89.5\% (V), 23.1\% (F) \\ 
                \hline
                \rowcolor[HTML]{FFFFFF} \cite{luo2017patient} &5 &Yes &Yes &Yes &DNN and Stacked Denoising Autoencoder (SDA) &No &No &Acc: 97,5\% N: Sen = 99,0\%, Ppv = 98,4\% S: Sen = 71,4\%, Ppv = 94,4\% V: Sen = 93,3\%, Ppv = 93,3\% F: Sen = 82,7\%, Ppv = 58,5\% \\ 
                \hline
                \rowcolor[HTML]{F2F2F2} \cite{hu2022transformer} &4, 8 and 2&Yes &No &Yes &CNN and Transformers &Yes &No &4 classes: Acc 99,49\%, F1-score 93,88\% 8 classes: Acc 99,09\%, F1-score 98,16\% 2 classes: Acc 99,23\%, F1-score 99,23\% \\ 
                \hline
                \rowcolor[HTML]{FFFFFF} \cite{raza2022designing} &5 &Yes &No &Yes &1D-CNN &Yes &No &Acc without noise: 98,9\% Acc with noise: 94,5\% Precisão: entre 89\% e 99\% dependendo da classe. Recall: entre 89\% e 99\% dependendo da classe. F1-Score: entre 91\% e 99\% dependendo da classe. \\ 
                \hline
                \rowcolor[HTML]{F2F2F2} \cite{ullah2021hybrid} & 5 and 8 &No &No &Yes &1D-CNN and 2D-CNN &No &No &1D-CNN: Acc de 97,38\%. 2D-CNN: Acc de 99,02\%. \\ 
                \hline
                \rowcolor[HTML]{FFFFFF} \cite{li2018deep} &5 &Yes &No &Yes &CNN &No &No &Acc: 98.9\% Prec: 95\% (N), 98\% (S), 97\% (V), 99\% (F), 100\%(Q) Recall: 99\% (N), 97\% (S), 99\% (V), 93\% (F), 100\%(Q) F1-Score: 97\% (N), 98\% (S), 98\% (V), 96\% (F), 100\%(Q) \\ 
                \hline
                \rowcolor[HTML]{F2F2F2} \cite{amirshahi2019ecg} &4 &Yes &No &Yes &Spiking Neural Networks - SNN &Yes &No &Acc: 97.9\% Sen: 80.2\% Spe: 99.8\% Prec: 97.3\% F1-score: 88.0\% \\ 
                \hline
                \rowcolor[HTML]{FFFFFF} \cite{yildirim2020accurate} &7 &No &Yes &No &CNN and LSTM &No &No &AFIB: Sen: 96,17\%, Prec: 94,16\%, Spec: 98,30\%, F-Score: 95,15\%, Acc: 97,82\%. GSVT: Sen: 89,94\%, Prec: 96,75\%, Spec: 99,30\%, F-Score: 93,22\%, Acc: 97,54\%. SB: Sen: 98,75\%, Prec: 98,25\%, Spe: 98,93\%, F-Score: 98,50\%, Acc: 98,86\%. SR: Sen: 96,88\%, Prec: 93,96\%, Spe: 98,32\%, F-Score: 95,40\%, Acc: 98,01\%. Acc: 96,13\% \\ 
                \hline
                \rowcolor[HTML]{F2F2F2} \cite{wang2021automated} &5 &Yes &No &Yes &CNN and Non-local Convolutional Block Attention Module (NCBAM) &No &No &Acc: 98,64\% Sen: 96,60\% Spe: 99,15\% F1-score: 96,64\% \\ 
                \hline
                \rowcolor[HTML]{FFFFFF} \cite{golrizkhatami2018ecg} &5 &Yes &Yes &Yes &CNN and SVM &No &No &Acc: 98\% Sen: 99.4\% (N), 75.6\% (S), 96.8\% (V), 93.8\% (F) Ppr: 98.6\% (N), 96.8\% (S), 95.1\% (V), 65.7\% (F) \\ 
                \hline
            \end{tabularx}
        \end{table}

        \begin{table}[!htp]
            \centering
            \caption{Comparative summary of studies on arrhythmia classification: characteristics of analyzed articles, including adherence to AAMI recommendations, use of the inter-patient paradigm, database used, classifier model, embedded system feasibility, on-chip learning capability, and reported performance metrics (Part 3 of 5).}
            \label{tab:table3}
            \scriptsize
            \renewcommand{\arraystretch}{1} 
            \begin{tabularx}{\linewidth}{|>{\centering\arraybackslash}p{0.6cm}|>{\centering\arraybackslash}p{0.8cm}|>{\centering\arraybackslash}p{0.8cm}|>{\centering\arraybackslash}p{0.8cm}|>{\centering\arraybackslash}p{0.8cm}|>{\centering\arraybackslash}p{2cm}|>{\centering\arraybackslash}p{1cm}|>{\centering\arraybackslash}p{1cm}|X|}
                \hline
                \rowcolor[HTML]{CCCCCC}
                \textbf{Article} & \textbf{Classes} & \textbf{Follow AAMI} & \textbf{Inter-patient} & \textbf{MIT-BIH} & \centering \textbf{Classifier} & \textbf{Embedded} & \textbf{On-Chip Learning} & \textbf{Performance} \\ 
                \hline
                \rowcolor[HTML]{F2F2F2} \cite{teijeiro2016heartbeat} &5 &Yes &Yes &Yes &Abductive rule-based interpretation and QRS clustering &No &No &Sen: 92,82\% (V), 85,10\% (S) Ppr: 92,23\% (V), 84,51\% (S) \\ 
                \hline
                \rowcolor[HTML]{FFFFFF} \cite{guo2019inter} &5 &Yes &Yes &Yes &DenseNet and Gated Recurrent Unit (GRU) &No &No &F1-Score: 61,94\% (S), 89,75\% (V) \\ 
                \hline
                \rowcolor[HTML]{F2F2F2} \cite{he2020framework} &5 &Yes &Yes &Yes &Dynamic Heartbeat Classification with Adjusted Features (DHCAF) and Multi-channel Heartbeat Convolution Neural Network (MCHCNN) &No &No &DHCAF: Acc: 91,4\% Sen: 84,0\% (S), 93,6\% (V) Ppv: 40,6\% (S), 67,3\% (V) MCHCNN: Acc: 93,0\% Sen: 39,1\% (S), 90,1\% (V) Ppv: 42,3\% (S), 72,0\% (V) \\ 
                \hline
                \rowcolor[HTML]{FFFFFF} \cite{naz2021ecg} &2 &No &No &Yes &CNN and SVM &No &No &Acc: 97,6\% Sen: 97,12\% Spe: 95,99\% \\ 
                \hline
                \rowcolor[HTML]{F2F2F2} \cite{hong2019combining} &4 &No &No &No &DNN and XGBoost &No &No &F1-Score: 0.9117 (N), 0.8128 (A), 0.7505 (O), 0.5671 (P) \\ 
                \hline
                \rowcolor[HTML]{FFFFFF} \cite{ji2019electrocardiogram} &5 &Yes &No &Yes &Faster Regions with Convolutional Neural Network (Faster R-CNN) &No &No &Mean Acc: 99,21\% Sen: 98.27\% (N), 98.77\% (L), 97.65\% (R), 97.54\% (V), 98.07\% (F) Spe: 99.39\% (N), 99.47\% (L), 99.43\% (R), 99.44\% (V), 99.50\% (F) \\ 
                \hline
                \rowcolor[HTML]{F2F2F2} \cite{yan2021energy} &4 &Yes &Yes &Yes &CNN and Spiking Neural Network (SNN) &Yes &No &Mean Acc: 92\% Energy consumption: 8,94 W (for 2 classes) e 13,23 W (for 4 classes) Sen: 91\% (S), 85\% (V) \\ 
                \hline
                \rowcolor[HTML]{FFFFFF} \cite{mehari2022self} &5 &No &No &No &Contrastive Predictive Coding (CPC) and CNN &No &No &Acc: 92,72\% AUC: 94,18\% \\ 
                \hline
                \rowcolor[HTML]{F2F2F2} \cite{shi2020automated} &4 &Yes &Yes &Yes &CNN and LSTM &No &No &Acc: 94,20\% Sen: 95,26\% (N), 90,74\% (S), 92,92\% (V) \\ 
                \hline
                \rowcolor[HTML]{FFFFFF} \cite{houssein2022automatic} &5 &Yes &No &Yes &CNN and Marine Predators Algorithm (MPA) &No &No &Acc: 99,33\% Sen: 98,52\% Prec: 98,79\% F-Score: 98,65\% \\ 
                \hline
                \rowcolor[HTML]{F2F2F2} \cite{liu2022arrhythmia} &5 &No &No &Yes &Autoencoder and LSTM &No &No &Acc: 98.57\% Sen: 97.98\% Ppv: 97.55\% \\ 
                \hline
                \rowcolor[HTML]{FFFFFF} \cite{de2018robust} &5 &Yes &Yes &Yes &Optimum-Path Forest (OPF) &No &No &Acc: 91,21\% \\ 
                \hline
                \rowcolor[HTML]{F2F2F2} \cite{sun2021beatclass} &5 &Yes &Yes &Yes &Bidirectional LSTM (BiLSTM) and GAN (Generative Adversarial Network) &No &No &Acc: 98,7\% F1-score: 99,5\% (N), 94,7\%(S), 97,0\%(V) \\ 
                \hline
                \rowcolor[HTML]{FFFFFF} \cite{wang2020improved} &5 &Yes &No &Yes &CNN &No &No &Acc: 99,06\% Sen: 99,61\% (N), 70,22\% (S), 95,08\% (V), 63,59\% (F) \\ \hline
                \rowcolor[HTML]{F2F2F2} \cite{pandey2019automatic} &5 &Yes &No &Yes &CNN with 11 layers &No &No &Acc: 98,30\% Mean Prec: 86,06\% Mean Recall: 95,51\% Mean F1-Score: 89,87\% \\ 
                \hline
                \rowcolor[HTML]{FFFFFF} \cite{dias2021arrhythmia} &3 &Yes &Yes &Yes &Linear Discriminant (LD) &No &No &Sen: 94.5\% (N), 92.5\% (S), 88.6\% (V) Ppv: 99.4\% (N), 39.9\% (S), 94.6\% (V) \\ 
                \hline
                \rowcolor[HTML]{F2F2F2} \cite{chen2020multi} &5 &Yes &Yes &Yes &CNN and Bidirectional Long Short-Term Memory (BLSTM) &No &No &Acc: 96,77\% F1-Score: 77,83\% \\ 
                \hline
                \rowcolor[HTML]{FFFFFF} \cite{chen2017energy} &2 &No &No &Yes & Weak Linear Classifier (WLC)  with SVM &Yes &No &Acc: 98,2\% Sen: 98,2\% Spe: 98,1\% \\ \hline
                \rowcolor[HTML]{F2F2F2} \cite{zhai2020semi} &5 &Yes &Yes &Yes &2D - CNN &No &No &Acc: 96,4\% Sen: 75,6\% (S), 91,8\% (V) Ppv: 76,4\% (S), 76,2\% (V) F1-Score: 76,0\% (S), 83,3\% (V) \\ 
                \hline
                \rowcolor[HTML]{FFFFFF} \cite{zhang2023hybrid} &4 &No &No &No &CNN with SVM &No &No &Mean F1-Score: 84,3\% Mean Prec: 81,63\% Mean Sen: 87,80\% \\ 
                \hline
                \rowcolor[HTML]{F2F2F2} \cite{baygin2021automated} &7 &No &No &No &SVM and Homeomorphically Irreducible Tree (HIT) &Yes &No &Acc (7 classes): 92,95\% Acc (4 classes): 97,18\% \\ 
                \hline
                \rowcolor[HTML]{FFFFFF} \cite{wan2020heartbeat} &5 &No &No &Yes &1D-CNN &No &No &Acc: 99,10\% Normal (N): Sen = 99,2\%, Ppv = 98,0\% Ventricular Premature Beat: Sen = 97,7\%, Ppv = 99,0\% Right Bundle Branch Block: Sen = 99,6\%, Ppv = 98,4\% Left Bundle Branch Block: Sen = 99,0\%, Ppv = 99,1\% Paced Beat: Sen = 99,9\%, Ppv = 99,8\% \\ 
                \hline
                \rowcolor[HTML]{F2F2F2} \cite{meng2022enhancing} &3 &No &No &No &Lightweight Fussing Transformer &Yes &No &Class N: Prec: 0,9975; Sen: 0,9984; F1: 0,9979 Class S: Prec: 0,9386; Sen: 0,8300; F1: 0,8810 Class V: Prec: 0,9158; Sen: 0,9447; F1: 0,9300 Acc: 0,9932 \\ 
                \hline
            \end{tabularx}
        \end{table}

        \begin{table}[!htp]
            \centering
            \caption{Comparative summary of studies on arrhythmia classification: characteristics of analyzed articles, including adherence to AAMI recommendations, use of the inter-patient paradigm, database used, classifier model, embedded system feasibility, on-chip learning capability, and reported performance metrics (Part 4 of 5).}
            \label{tab:table4}
            \scriptsize
            \renewcommand{\arraystretch}{1.2}
            \begin{tabularx}{\linewidth}{|>{\centering\arraybackslash}p{0.6cm}|>{\centering\arraybackslash}p{0.8cm}|>{\centering\arraybackslash}p{0.8cm}|>{\centering\arraybackslash}p{0.8cm}|>{\centering\arraybackslash}p{0.8cm}|>{\centering\arraybackslash}p{2cm}|>{\centering\arraybackslash}p{1cm}|>{\centering\arraybackslash}p{1cm}|X|}
                \hline
                \rowcolor[HTML]{CCCCCC}
                \textbf{Article} & \textbf{Classes} & \textbf{Follow AAMI} & \textbf{Inter-patient} & \textbf{MIT-BIH} & \centering \textbf{Classifier} & \textbf{Embedded} & \textbf{On-Chip Learning} & \textbf{Performance} \\ 
                \hline
                \rowcolor[HTML]{FFFFFF} \cite{gupta2022novel} &Not specified &No &No &Yes &Fractional Wavelet Transform (FrWT) and Principal Component Analysis (PCA) &No &No &Acc: 99,89\% MSE: 0,1656\% \\ 
                \hline
                \rowcolor[HTML]{F2F2F2} \cite{hammad2022deep} &20 &No &No &Yes &CNN and LSTM &No &No &Acc (MIT-BIH): 97\% Acc (PhysioNet 2016): 94\% \\ 
                \hline
                \rowcolor[HTML]{FFFFFF} \cite{huang2022snippet} & 5 and 8 &No &No &No &SPN-V2 and Knee-Guided Neuroevolution Algorithm (KGNA) &No &No &PTBXL: Acc: 82,6\% F1-score: 78,8\% CPSC2018: Acc: 83,6\% F1-score: 78,7\% \\ 
                \hline
                \rowcolor[HTML]{F2F2F2} \cite{barandas2024evaluation} &9 &No &No &No &CNN and GRU &No &No &Acc: 0,903 F1-Score: 0,856 \\ 
                \hline
                \rowcolor[HTML]{FFFFFF} \cite{shekhawat2024binarized} &2 &No &No &No &Binarized Spiking Neural Network (BSNN) &No &No &Acc: 97,5\% Pre: 95,2\% Sen: 96,5\% Recall: 96,8\% Spe: 98,4\% \\
                \hline
                \rowcolor[HTML]{F2F2F2} \cite{ji2024msgformer} &9 &No &No &Yes &Multi-Scale Grid Transformer (MSGformer) &No &No &Acc: 99,28\% Sen: 97,13\% Ppv: 97,87\% \\ 
                \hline
                \rowcolor[HTML]{FFFFFF} \cite{alamatsaz2024lightweight} &9 &No &No &Yes &CNN and LSTM &Yes &No &Acc: 98.24\% Sen: 86.1\% Spe: 97.5\% \\ 
                \hline
                \rowcolor[HTML]{F2F2F2} \cite{anand2024enhanced} &5 &Yes &No &Yes &ResNet-50 &No &No &Acc: 98,14\% Sen: 97,3\% Ppv: 96,5\% \\ 
                \hline
                \rowcolor[HTML]{FFFFFF} \cite{tao2024ecg} & 5, 6 and 17 &No &No &No &Dual Kernel Residual Block (DKR-block) and 2D-CNN &No &No &PTB-XL: AUCmacro: 0,929, F1macro: 0,770 HFECGIC: AUCmacro: 0,981, F1micro: 0,932 LUDB: AUCmacro: 0,998, F1macro: 0,989 \\ 
                \hline
                \rowcolor[HTML]{F2F2F2} \cite{zubair2023deep} &5 &Yes &Yes &Yes &1D-CNN &No &No &Acc: 96.19\% Sen: 95,63\% (N), 96.30\% (S), 97.30\% (V) Spe: 86,59\% (N), 97.29\% (S), 96.30\% (V) Ppv: 98.42\% (N), 95.31\% (S), 97.88\% (V) \\ 
                \hline
                \rowcolor[HTML]{FFFFFF} \cite{khatar2024advanced} &2 &No &No &Yes &Inception-ResNet and BiLSTM &No &No &Acc: 98,94\% Prec: 97,20\% Sen: 94,08\% F1-Score: 95,58\% AUC-ROC: 94,08\% \\ 
                \hline
                \rowcolor[HTML]{F2F2F2} \cite{chen2024novel} &5 &Yes &No &Yes &CNN, autoencoder and 1D-CBAM &No &No &F1-Score: 98,73\% Acc: 99,61\% \\ 
                \hline
                \rowcolor[HTML]{FFFFFF} \cite{sun2022scalable} &3 &Yes &No &Yes &SCALT &Yes &No &Prec: 98,65\% \\ 
                \hline
                \rowcolor[HTML]{F2F2F2} \cite{ganeshkumar2021explainable} &8 &No &No &No &CNN &No &No &Acc: 96,2\%. Prec: 0,986. Recall: 0,949. F1-Score: 0,967 \\ 
                \hline
                \rowcolor[HTML]{FFFFFF} \cite{xia2023generative} &5 &Yes &Yes &Yes &CNN and Bi-LSTM &No &No &Acc: 94,69\% F1-Score: 62,02\% (S), 85,71\% (V) \\ 
                \hline
                \rowcolor[HTML]{F2F2F2} \cite{yang2023multi} & 9, 34 and 71 &No &No &No &CNN &No &No &AUC: 92.89 Sen: 73.66 \\ 
                \hline
                \rowcolor[HTML]{FFFFFF} \cite{zahid2022global} &3 &Yes &Yes &Yes &Self-Operational Neural Networks (Self-ONN) &No &No &Prec: 99.21\% (N), 82.19\% (S), 94.41\% (V) Recall: 99.10\% (N) , 82.50\% (S), 96.10\% (V) F1-score: 99.15\% (N), 82.34\% (S), 95.2\% (V) \\ 
                \hline
                \rowcolor[HTML]{F2F2F2} \cite{kumar2023fuzz} &5 &Yes &No &Yes &CNN &No &No &MIT-BIH Arrhythmia Dataset: Acc: 98,66\% Prec: 98,92\% Recall: 93,88\% F1-Score: 96,34\% PTB Diagnostic ECG Dataset: Acc: 95,79\% Prec: 96,29\% Recall: 85,38\% F1-Score: 80,37\% \\ 
                \hline
                \rowcolor[HTML]{FFFFFF} \cite{rawal2023hardware} &4 &No &No &No &1D-CNN &Yes &No &Acc 96,77\% (software) and 86,37\% (hardware FPGA) Acc for Atrial Fibrillation (AF): 99.17\% (software), 97.34\% (hardware) F1-Score for Atrial Fibrillation (AF): 95.34\% (software), 86.92\% (hardware). \\ 
                \hline
                \rowcolor[HTML]{F2F2F2} \cite{han2023multimodal} &6 and 7 &Yes &Yes &Yes &Multimodal Neural Network with Multi-Instance Learning (MAMIL) &No &No &St. Petersburg INCART Arrhythmia Dataset: AUC: 74,95\% F1: 69,10\% Recall: 67,89\% MIT-BIH Supraventricular Arrhythmia Dataset: AUC: 79,45\% F1: 60,99\% Recall: 66,08\% \\ 
                \hline
                \rowcolor[HTML]{FFFFFF} \cite{farag2023tiny} &5 &Yes &Yes &Yes &CNN &Yes &No &Acc: 98,18\% F1-Score: 92,17\% Sen: 85,30\% (S), 96,34\% (V) \\ 
                \hline
                \rowcolor[HTML]{F2F2F2} \cite{le2023lightx3ecg} &3 &No &No &No &1D-SEResNet with a Lead-wise Attention Module &Yes &No &Chapman: Mean F1-Score : 0.9718 Mean Prec: 0.9736 Mean Recall: 0.9703 Mean Acc: 0.9873 CPSC-2018 F1-Score: 0.8004 Mean Prec: 0.8209 Mean Recall: 0.7862 Mean Acc: 0.9628 \\ 
                \hline
                \rowcolor[HTML]{FFFFFF} \cite{jin2021novel} &5 &No &No &Yes &DLA-CLSTM - CNN and BiLSTM &No &No &Acc: 88,76\% F1-Macro: 80,54\% \\ 
                \hline
                \rowcolor[HTML]{F2F2F2} \cite{pokaprakarn2021sequence} &5 &No &No &Yes &CNN and RNN &No &No &Acc: 97,60\% Mean F1-Score: 0,89 \\ 
                \hline
                \rowcolor[HTML]{FFFFFF} \cite{degirmenci2022arrhythmic} &5 &No &No &Yes &2D-CNN &No &No &Acc: 99,7\% Sen: 99,7\% Spe: 99,22\% \\ 
                \hline
                \rowcolor[HTML]{F2F2F2} \cite{malik2021real} &5 &Yes &No &Yes &1D Self-ONN &No &No &Acc: 98\% (S), 99,04\% (V) F1-Score: 76,6\% (S), 93,7\% (V) \\ 
                \hline
            \end{tabularx}
        \end{table}

        \begin{table}[!htp]
            \centering
            \caption{Comparative summary of studies on arrhythmia classification: characteristics of analyzed articles, including adherence to AAMI recommendations, use of the inter-patient paradigm, database used, classifier model, embedded system feasibility, on-chip learning capability, and reported performance metrics (Part 5 of 5).}
            \label{tab:table5}
            \scriptsize
            \renewcommand{\arraystretch}{1.2}
            \begin{tabularx}{\linewidth}{|>{\centering\arraybackslash}p{0.6cm}|>{\centering\arraybackslash}p{0.8cm}|>{\centering\arraybackslash}p{0.8cm}|>{\centering\arraybackslash}p{0.8cm}|>{\centering\arraybackslash}p{0.8cm}|>{\centering\arraybackslash}p{2cm}|>{\centering\arraybackslash}p{1cm}|>{\centering\arraybackslash}p{1cm}|X|}
                \hline
                \rowcolor[HTML]{CCCCCC}
                \textbf{Article} & \textbf{Classes} & \textbf{Follow AAMI} & \textbf{Inter-patient} & \textbf{MIT-BIH} & \centering \textbf{Classifier} & \textbf{Embedded} & \textbf{On-Chip Learning} & \textbf{Performance} \\ 
                \hline
                \rowcolor[HTML]{FFFFFF} \cite{bing2022electrocardiogram} &5 &Yes &No &Yes &Convolutional Vision Transformer (ConViT) &No &No &Acc: 99,5\% Spec: 99,9\% (S), 99,7\% (V) F1-Score: 95,0\% (S), 97,7\% (V) \\ 
                \hline
                \rowcolor[HTML]{F2F2F2} \cite{ribeiro2022ecg} &5 &Yes &No &Yes &Quantized 1D-CNN &Yes &No &Acc: 99,6\% Sen: 100\% (V) Spe: 99,9\% (V) F1-Score: 99,9\% (V) \\ 
                \hline
                \rowcolor[HTML]{FFFFFF} \cite{feng2022unsupervised} &5 &Yes &Yes &Yes &Unsupervised Semantic-Aware Adaptive Feature Fusion Network (USAFFN) &No &No &Acc: 95,7\% F1-Score: 89,4\% (V), 75,7\% (S), 43,1\% (F) \\ 
                \hline
                \rowcolor[HTML]{F2F2F2} \cite{chu2022neuromorphic} &5 &Yes &No &Yes & SNN and recurrent MLP (rMLP) &Yes &No &Acc: 98,22\%. \\ 
                \hline
                \rowcolor[HTML]{FFFFFF} \cite{ran2022homecare} &26 &No &No &No &DCNN &Yes &No &Mean F1-Score: 0,913. Mean AUC: 0,944. Mean Sen: 0,891. Mean Spe: 0,997. \\ 
                \hline
                \rowcolor[HTML]{F2F2F2} \cite{tesfai2022lightweight} &4 &Yes &No &No &1D-CNN &Yes &No &Acc: 92,23\% F1-score: 97\% \\ 
                \hline
                \rowcolor[HTML]{FFFFFF} \cite{shao2020wearable} &4 &No &No &Yes &CatBoost &No &No &F1-Score: 0,92 Prec: 96\% Sen for AF: 90\% Spe for AF: 99,61\% \\ 
                \hline
                \rowcolor[HTML]{F2F2F2} \cite{yin20212} &2 &No &No &Yes &Derivative-Based Adaptive Arrhythmia Detection Algorithm &Yes &No &Ppv: 99,3\%. Sen: 98,2\%. \\ 
                \hline
                \rowcolor[HTML]{FFFFFF} \cite{raj2018personalized} &5 &Yes &Yes &Yes &Least-Square Twin Support Vector Machine (LSTSVM) & Yes &No &Acc: 96,29\% Sen: 96,08\% F1-Score: 96,08\% \\ 
                \hline
                \rowcolor[HTML]{F2F2F2} \cite{tsoutsouras2017exploration} &2 &No &No &Yes &SVM &Yes &No &Sen and Spe: > 98\% \\ 
                \hline
                \rowcolor[HTML]{FFFFFF} \cite{sakib2021proof} &4 &Yes &No &Yes &1D-CNN &Yes &No &Acc: 95.83\% \\ 
                \hline
                \rowcolor[HTML]{F2F2F2} \cite{rincon2020iot} &4 &No &No &No &MobileNet &Yes &No &Acc for Atrial Fibrillation (AF): 90\% Acc for Normal Sinus Rhythm (NSR): 89\% Acc for Very Noisy Signals (NO): 92\% Acc for Other Rhythms (OR): 95\% \\ 
                \hline
                \rowcolor[HTML]{FFFFFF} \cite{obeidat2021hybrid} &6 &Yes &No &Yes &CNN-LSTM &No &No &Acc: 98,22\% Sen: 98,23\% Spe: 99,64\% Prec: 98,26\% \\ 
                \hline
                \rowcolor[HTML]{F2F2F2} \cite{lu2021efficient} &5 &No &No &Yes &1D-CNN &Yes &No &Acc: 99,10\% \\ 
                \hline
                \rowcolor[HTML]{FFFFFF} \cite{jeon2020lightweight} &5 &Yes &No &Yes &RNN &Yes &No &Acc: 99,80\% \\ 
                \hline
                \rowcolor[HTML]{F2F2F2} \cite{xing2022accurate} &5 &Yes &Yes &Yes & SNN &Yes &No &Acc: 98,26\% Sen: 94,75\% F1-Score: 89,09\% \\ 
                \hline
                \rowcolor[HTML]{FFFFFF} \cite{wu2019neural} &5 &Yes &No &Yes &BLSTM &Yes &No & Acc: 98.3\% (S), 98.6\% (V). \newline Sen: 80.6\% (S), 91.8\% (V). \newline Ppr: 82.4\% (S), 94.0\% (V) \\ 
                \hline
                \rowcolor[HTML]{F2F2F2} \cite{lu2022efficient} &5 &No &No &No &Multiple Receptive Field Convolutional Neural Network (MRF-CNN) &Yes &No &Acc: 98.6\% (V), 98.3\% (S) Sen: 91.8\% (V), 80.6\% (S) Spe: 99.4\% (V), 99.1\% (S) Prec: 94.0\% (V), 82.4\% (S) F1-Score: 92.9\% (V), 81.5\% (S) \\ \hline
                \rowcolor[HTML]{FFFFFF} \cite{feyisa2022lightweight} &5 &No &No &No &Lightweight MRF-CNN &No &No &F1-score: 0.72 AUC: 0.93. \\ 
                \hline
                \rowcolor[HTML]{F2F2F2} \cite{cheng2022efficient} &5 &No &No &Yes &1D U-net with 22 layers &Yes &No &Acc: 95,55\% Ppr: 95,52\% Sen: 95,55\% Spe: 97,64\% \\ 
                \hline
                \rowcolor[HTML]{FFFFFF} \cite{zhu2021robust} &5 &Yes &Yes &Yes &XGBoos &No &No &Acc: 99,14\% Sen: 97,4\% Ppv: 97,9\% F1-Score: 97,6\% \\ 
                \hline
                \rowcolor[HTML]{F2F2F2} \cite{mao2022ultra} &5 &Yes &Yes &Yes &SNN and ANN &Yes &Yes &Acc: 97,36\%. Spe: 82,30\%. F1-Score: 98,66\% \\
                \hline
                \rowcolor[HTML]{FFFFFF} \cite{xu20222} &5 &Yes &No &Yes &Cardiac Arrhythmia Watchdog (CAW) &Yes &No &Sen: 97,62\% Spe: 97,23\% Acc: 97,58\% \\ 
                \hline
                \rowcolor[HTML]{F2F2F2} \cite{deng2023energy} &2 &No &No &Yes &Linear Pre-Classifier (LPC) and Non-Linear Support Vector Machine (NLSVM) &Yes &No &Acc: 97,34\% Sen: 99,83\% Ppv: 99,65\% \\
                \hline
                \rowcolor[HTML]{FFFFFF} \cite{berrahou2024arrhythmia} &5 &Yes &Yes &Yes &1D-CNN &No &No &Acc: 98.73\% Sen: 99.05\% (N), 79.02\% (S), 96.79\% (V) Prec: 99.73\% (N), 95.02\% (S), 96.33\% (V) \\ 
                \hline
                \rowcolor[HTML]{F2F2F2} \cite{zhang2023configurable} &4 &No &No &Yes &1D-CNN &Yes &No &Acc: 98.9\% Sen: 98.9\% \\ 
                \hline
                \rowcolor[HTML]{FFFFFF} \cite{zhang2023low} &5 &No &No &Yes &MLP &Yes &No &Acc: 96,69\% Prec: 96,67\% \\ 
                \hline
                \rowcolor[HTML]{F2F2F2} \cite{fatimah2024ecg} &5 &Yes &Yes &Yes &SVM &No &No &Acc: 98,03\%. F1-scores: 99,46\% (N), 89,35\% (S), 89,62\% (V) \\ 
                \hline
                \rowcolor[HTML]{FFFFFF} \cite{yazid2024atrial} &2 &No &No &Yes &SVM &Yes &No &Acc: 99,47\% Sen: 99,49\% Spe: 99,46\% \\ 
                \hline
            \end{tabularx}
        \end{table}
    
    \subsection{Influence of ECG Classification Studies: Citation Network Analysis}

        Understanding which studies have most influenced the field of ECG classification is essential for mapping the evolution of methodologies and identifying research gaps. This section presents a citation network graph that visually represents the most impactful works in this domain, illustrating how knowledge has been built and disseminated over time. By analyzing citation relationships, we can highlight contributions, track the adoption of AAMI and inter-patient standards, and assess which studies have shaped contemporary research directions.

\begin{sidewaysfigure*}
            \centering
            \includegraphics[width=\linewidth,trim={10cm 0 10cm 0},clip]{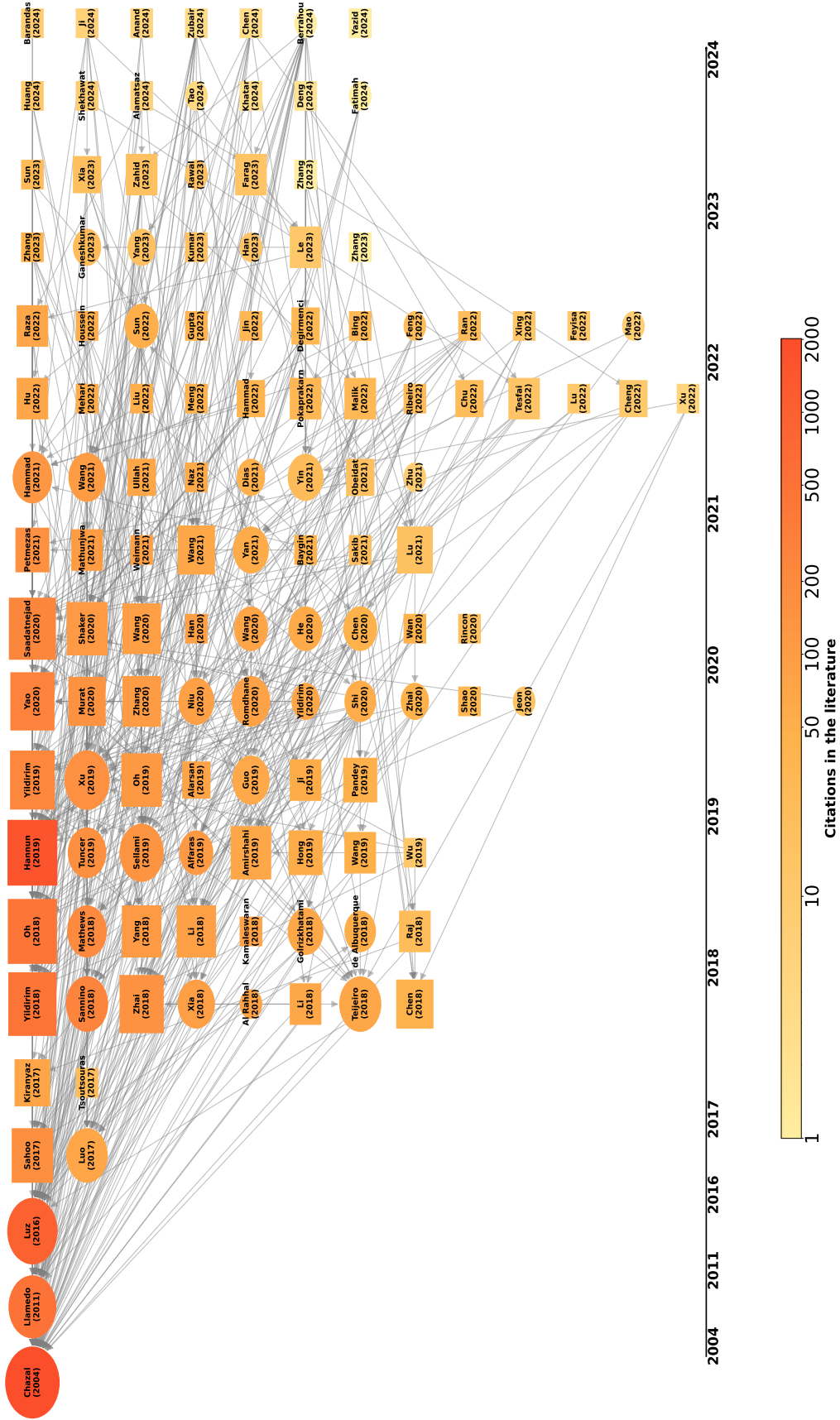}
            \caption{Citation network of the reviewed studies in ECG arrhythmia classification. Each node represents an individual study, while edges indicate citation relationships. The direction of the arrows shows the citation flow, where the source node (origin of the arrow) is the citing article, and the target node (end of the arrow) is the cited article. The color of the nodes represents the number of times a study has been cited in the literature, with darker shades indicating higher citation counts. The shape of the nodes distinguishes whether the study followed the inter-patient paradigm proposed by De Chazal: round nodes indicate studies that implemented this method, while rectangular nodes represent studies that did not follow this approach.}
            \label{fig:graph}
        \end{sidewaysfigure*}

        Figure \ref{fig:graph} presents the citation graph, which consists of nodes representing individual articles and directional edges (arrows) indicating citation relationships, where an arrow originates from the citing article and points to the cited one. The size of each node is logarithmically proportional to the number of citations received within the 122 articles analyzed in this review, meaning that larger nodes correspond to studies that have been more frequently referenced within this set of studies. Furthermore, the color of each node reflects the total number of citations in the broader scientific literature, as retrieved from Web of Science as of August 2024. Warmer colors indicate a higher overall citation count, emphasizing the global academic influence of each study. Additionally, the shape of each node distinguishes whether the study followed the inter-patient paradigm (round nodes) or not (rectangular nodes). The code used to generate the graph in Figure \ref{fig:graph} is available in the following GitHub repository, which also provides the necessary steps for reproduction: \url{https://github.com/silvagal/Plot-Citation-Graph.git}.

        To provide a more comprehensive context, a few influential articles outside the primary review set have been included in the analysis. Luz \textit{et al.} \cite{luz_ecg-based_2016} is an important survey on ECG classification that serves as the foundation for this review, helping to identify gaps in the literature over time. De Chazal \textit{et al.} \cite{de_chazal_automatic_2004} is featured as the work that introduced the inter-patient paradigm. Lastly, Llamedo \textit{et al.} \cite{llamedo2010heartbeat} is highlighted as a study that not only rigorously applied the inter-patient paradigm but also validated its generalization across multiple databases, including MIT-BIH Arrhythmia, MIT-BIH Supraventricular, and INCART.

        Analyzing the citation graph in Figure \ref{fig:graph}, we can see that square nodes, representing studies that did not adopt the De Chazal's methodology, were significantly more influential within the reviewed articles than round nodes. This trend is particularly pronounced in the years 2017, 2018, 2019, 2022, and 2023, where larger square nodes outnumber their round counterparts in terms of size and prominence.

        The peak influence of studies that did not implement the inter-patient methodology 
        occurred between 2018, 2019 and 2020 with the most notable contributions coming from Yıldırım \textit{et al.} \cite{yildirim2018arrhythmia} , Hannun \textit{et al.} \cite{hannun2019cardiologist}, and Saadatnejad \textit{et al.} \cite{saadatnejad2019lstm}. These works had substantial impact within the field, as indicated by their large node sizes and dark coloration in the graph, signifying high citation counts.

        The approach proposed by Yıldırım \textit{et al.} \cite{yildirim2018arrhythmia}, was a deep learning-based approach using a 1D-Convolutional Neural Network (1D-CNN) to classify 17 types of cardiac arrhythmias from long-duration ECG signals. Their model was trained and evaluated using the MIT-BIH Arrhythmia Database, specifically 1,000 randomly selected 10-second ECG fragments recorded from 45 individuals. Unlike traditional methods that focused on detecting individual QRS complexes, this study used full 10-second ECG segments for classification. The 16-layer deep CNN model was designed to operate in an end-to-end fashion, eliminating the need for handcrafted feature extraction.

        The results demonstrated high accuracy, with an overall classification performance of 91.33\% for 17 classes, 92.51\% for 15 classes, and 95.2\% for 13 classes. The model achieved a classification time of just 0.015 seconds per ECG segment, making it highly efficient for real-time applications. However, while the study utilized the MIT-BIH dataset, it did not adhere to the E3C, as it neither implemented AAMI recommendations nor applied the inter-patient division method. This limitation not only impacted the reproducibility of results in real-world scenarios, where classification models had to handle previously unseen heartbeats, but also made it difficult to compare the model’s performance with other benchmarks designed for real-world generalization.

        Furthermore, although the authors discussed the possibility of implementing the model in mobile and cloud-based systems due to its low computational complexity, the study did not explicitly evaluate the feasibility of embedding the model into low-power hardware.

        The work of Hannun \textit{et al.} \cite{hannun2019cardiologist} proposed a deep learning model trained end-to-end to classify 12 types of cardiac rhythms using a large private dataset of 91,232 ECG recordings from 53,549 patients, collected via a single-lead ambulatory monitor. The model was validated on an independent test set of 328 ECGs, achieving an impressive average AUC above 0.97 and an F1-score of 0.837, surpassing the average performance of expert cardiologists of 0.780. Additionally, it was evaluated on the PhysioNet Challenge 2017 dataset \cite{clifford2017af}, where it demonstrated results comparable to the best-performing models in the competition.
        

        Despite its strong results, the study did not adhere to any of the E3C standards, limiting the comparability of its findings with other benchmarks in the literature and restricting its applicability for continuous ECG monitoring in resource-constrained environments.
        
        Nevertheless, Hannun's study is one of the most influential works, not only within our review but also in the broader literature. Its representation as a large node underscores its prominence, while its square shape indicates that it did not follow the inter-patient paradigm.  While its methodology is robust and its results are excellent, the use of a private dataset restricts the reproducibility of the experiment, making it difficult for other researchers to validate its findings and limiting its role as a benchmark for future studies.

        Finally, The study by Saadatnejad \textit{et al.} \cite{saadatnejad2019lstm} proposed an ECG classification algorithm optimized for continuous cardiac monitoring on wearable devices with limited processing capacity. Their method employed a hybrid approach that combined wavelet transform with multiple Long Short-Term Memory (LSTM) recurrent neural networks (RNNs) to capture both temporal dependencies and signal features. Unlike computationally intensive deep learning models, this approach prioritized computational efficiency, ensuring real-time classification on embedded platforms.

        For training and evaluation, the authors used the MIT-BIH Arrhythmia Database, excluding records with paced beats (102, 104, 107, and 217) in accordance with AAMI standards. The model classified heartbeats into seven arrhythmia categories while also providing results in a five-class format for comparison with previous studies. Despite using the same test dataset proposed by De Chazal \textit{et al.} \cite{de_chazal_automatic_2004}, their training process did not adhere to the inter-patient partitioning methodology. Instead of strictly separating patients between training and testing sets, Saadatnejad \textit{et al.} \cite{saadatnejad2019lstm} included heartbeats in their training set that were designated for testing. Consequently, their model was exposed to heartbeats from patients in the test set during training, whereas De Chazal’s model, to which they compared their results, was evaluated solely on unseen patient data. This discrepancy makes the performance comparison between the two approaches not entirely equivalent.
        
        The model demonstrated superior classification performance compared to previous works. For VEB detection, accuracy exceeded 99\%, with an F1-score improvement of up to 19.8\% over prior methods. Similarly, for the class VEB, accuracy gains reached 3.1\%, with a 24.5\% improvement in F1-score. Importantly, the algorithm proved feasible for real-time execution on resource-constrained hardware, such as the Moto 360, NanoPi Neo Plus2, and Raspberry Pi Zero, achieving execution times between 31.2 ms and 58.6 ms per heartbeat.

        On the other hand, studies that followed the inter-patient methodology gained greater influence in 2021, particularly in the work of Hammad \textit{et al.} \cite{hammad2020multitier}. It is important to highlight other influential works from 2018 and 2019 that, while not the most impactful of those years, still had a significant presence in the literature due to their citation count and the number of times they were referenced in the articles included in this review. Notably, the studies by Sanino and De Pietro \cite{sannino2018deep} and Xu \textit{et al.} \cite{xu2018towards} contributed to this broader academic influence. These studies represent the most cited among those that adhered to the inter-patient paradigm.

        The study by Hammad \textit{et al.} \cite{hammad2020multitier} proposed a multi-tier deep learning model for arrhythmia detection, integrating deep learning, machine learning, and genetic algorithm (GA)-based optimization to enhance classification accuracy. The methodology consisted of three key components: a ResNet-LSTM-based deep learning model for feature extraction, GA for feature selection and optimization, and multiple classifiers, including k-nearest neighbors (k-NN), SVM, and MLP, for final classification.

        The dataset used was the MIT-BIH Arrhythmia Database, following the Association for the AAMI standards. The study categorized heartbeats into five standard arrhythmia classes (N, S, V, F, and Q) and conducted both intra-patient and inter-patient evaluations. For intra-patient validation, beats from all recordings were shuffled and split for five-fold cross-validation. For inter-patient validation, data from 22 patients were used for training, while another 22 were used for testing, following AAMI guidelines.
        
        The proposed model demonstrated high classification performance. In intra-patient evaluation, k-NN with GA achieved an accuracy of 98.0\%, specificity of 98.9\%, and sensitivity of 99.7\%. In inter-patient validation, which better simulates real-world deployment, the model maintained strong performance, outperforming many state-of-the-art approaches.
        
        Sannino and De Pietro proposed a deep learning-based ECG heartbeat classification model for arrhythmia detection. Their approach utilized a DNN with seven hidden layers, developed using the TensorFlow framework and implemented in Python. The model was designed empirically, adjusting the number of hidden layers, activation functions, and neurons per layer to optimize classification performance.

        The authors trained and evaluated their model on the MIT-BIH Arrhythmia Database. They followed the AAMI recommendations by categorizing heartbeats into five standard arrhythmia classes (N, S, V, F, and Q). Besides, the study followed the inter-patient partitioning.
        
        The experimental results demonstrated strong classification performance. The proposed DNN achieved an accuracy of 99.68\%, sensitivity of 99.48\%, and specificity of 99.83\% across the entire dataset, outperforming traditional machine learning classifiers such as SVM, KNN, and decision tree-based models. However, the study did not explore the feasibility of deploying the model on resource-constrained embedded systems, leaving its real-time applicability for wearable ECG monitoring devices unverified.

        Finally, the work by Xu \textit{et al.} \cite{xu2018towards} proposed an end-to-end ECG classification system using raw signal extraction and DNNs for automated heartbeat classification. Unlike traditional methods that rely on handcrafted feature extraction, their approach leveraged a DNN to perform both feature learning and classification, enhancing the representation of ECG signals. The model was designed to process time-domain ECG signals, extracting sample points using a sliding window approach that included not just the QRS complex but also the P and T waves, ensuring a more comprehensive feature set.

        The study was conducted using the MIT-BIH Arrhythmia Database, with a focus on a two-lead ECG configuration commonly used in long-term monitoring. The dataset was preprocessed according to the AAMI standard, mapping the arrhythmia classes into five major categories. The evaluation followed a subject-oriented partitioning scheme to ensure that training and testing were performed on different patients, enhancing the model’s generalization.
        
        The authors implemented a three-layer deep neural network, trained using 22-fold cross-validation and tested on unseen patient data. The model achieved an overall accuracy of 94.7\%, outperforming previous patient-independent classifiers. Specifically, it demonstrated higher sensitivity and specificity for Class S and Class V. The study also included an analysis of ROC curves, showing that the proposed method achieved superior classification performance across multiple sensitivity-specificity trade-offs.

        Although the study followed AAMI recommendations regarding dataset selection and class grouping, it did not explicitly evaluate the feasibility of deploying the model on embedded systems, thereby not adhering to the E3C.
        
        In this context, the overall trend suggests that studies not following the inter-patient methodology continued to dominate in terms of academic impact, raising questions about the extent to which real-world generalization has been prioritized in highly influential research.

    \subsection{Analysis of State-of-the-Art ECG Classification Models Meeting E3C}

        Our primary focus is to analyze the articles that fully adhered to the E3C. These articles form the core of this research investigation, as they align with the objective of approximating real-world scenarios where data are imbalanced, unseen heartbeats must be classified, and computational resources are severely constrained. According to Figure \ref{fig:criteria_analysis}, these studies account for only five articles (4.1\%), highlighting the limited attention these critical conditions have received in the literature. Therefore, we will analyze these five studies in detail to identify state-of-the-art methods that meet all four criteria.
    
        The first study to be analyzed is the one by Yan \textit{et al.} \cite{yan2021energy}, who presented a method for improving energy efficiency in ECG classification on embedded devices. The proposed approach combined convolutional neural networks (CNNs) and spiking neural networks (SNNs) in a two-stage classification system. In the first stage, a two-class CNN classified heartbeats as either normal or abnormal. Abnormal heartbeats were then passed to a second stage, where a four-class CNN refined the classification into four categories: Normal (N), Supraventricular Ectopic Beat (SVEB), Ventricular Ectopic Beat (VEB), and Fusion (F). To further reduce energy consumption, the authors converted the trained CNNs into SNNs, which use binary spikes for data transmission instead of real numbers. This conversion replaced computationally expensive multiplications with simpler addition operations, significantly lowering energy requirements.
    
        To address class imbalance in the dataset, the authors applied the Synthetic Minority Oversampling Technique (SMOTE), which augmented the minority classes and created a balanced training dataset. The training data were segmented into windows of 180 points centered on the R-peak, as annotated in the MIT-BIH dataset.
        
        The CNNs were implemented and trained using TensorFlow, with the two-class CNN trained for 10 epochs using backpropagation. The SNNs were created by transferring the trained weights from the CNNs and normalizing them to accommodate spike-based representation. Input data were encoded into spike trains, and thresholds for neuron activation were optimized to balance accuracy and energy efficiency. The energy consumption of the CNNs was measured on a system with an Intel Core i5 processor and NVIDIA GTX 1060 GPU, while the SNNs were evaluated using a custom neuromorphic accelerator, Shenjing.
        
        The results demonstrated that the two-class CNN achieved a sensitivity of 80\% and a positive predictive value of 98\% for normal beats, while abnormal beats attained a sensitivity of 87\% and a positive predictive value of 35\%, resulting in an average accuracy of 81\%. The four-class CNN further improved classification performance, achieving sensitivities of 91\% for SVEB and 85\% for VEB.
                
        Similarly, the two-class SNN achieved a sensitivity of 78\% and a positive predictive value of 98\% for normal beats, with abnormal beats achieving a sensitivity of 85\% and a positive predictive value of 35\%, maintaining an average accuracy of 81\%. Despite a slight 2\% drop in accuracy compared to the CNN, the SNN exhibited highly competitive performance. The four-class SNN achieved comparable sensitivities of 91\% for SVEB and 85\% for VEB, with minimal accuracy loss relative to the CNN.
        
        What sets the SNN apart is its energy efficiency, consuming only 0.077 W compared to the CNN's 10.40 W. This significant reduction in energy consumption, coupled with its competitive classification performance, underscores the SNN's suitability for deployment on resource-constrained embedded systems, particularly for real-time and continuous ECG monitoring applications.
    
        Farag \cite{farag2023tiny} proposed a methodology for ECG classification optimized for deployment on resource-constrained edge devices. The methodology was grounded in the matched filter (MF) interpretation of convolutional neural networks (CNNs), where the convolutional kernels were pre-assigned to represent class-specific templates of ECG heartbeats. This approach reduced the model complexity by eliminating the need to learn the convolutional kernel weights during training.
    
        The CNN architecture consisted of a single Conv1D layer with 13 filters (each corresponding to an ECG class in the MIT-BIH dataset), followed by Batch Normalization (BN) and Global Max Pooling (GMP) layers. Dynamic normalized RR intervals were used as auxiliary input features, processed by fully connected (FC) layers. The outputs of the GMP layer and RR intervals were concatenated and passed through a Softmax output layer to generate class probabilities. The input to the model could be either the raw ECG signal or its first derivative, with the latter demonstrating superior discriminative capabilities. The Conv1D kernels were initialized using the average of all heartbeat segments for each ECG class in the training dataset. This reduced the number of trainable parameters, accelerated training, and simplified the model.
        
        The methodology leveraged the inter-patient division of the MIT-BIH Arrhythmia Database to train and evaluate the model, ensuring realistic testing conditions by separating the training and testing sets by patient records. ECG signals were segmented into 0.5-second windows centered around R-peaks, and dynamic normalized RR intervals were calculated using a moving average over past intervals. The normalized RR intervals were essential for reducing inter-individual variations and enhancing the model’s generalization capability.
        
        Experiments demonstrated the efficacy of the proposed model both in cloud environments and on edge devices. On the cloud, the model achieved an average accuracy of 98.18\%, with sensitivities of 85.3\% for SVEB beats and 96.34\% for VEB beats. The F1-score for the entire model was 92.17\%. The use of the matched filter interpretation and the pre-defined kernel weights facilitated quick and stable training, with negligible overfitting.
    
        Comparing the proposed matched filter-based CNN model with state-of-the-art methods for inter-patient ECG classification,
        the proposed model achieved an accuracy of 98.18\%, sensitivities of 81.60\% for SVEB and 95.00\% for VEB, and an F1-score of 92.17\%, outperforming or matching the best results in the field. For example, it surpassed the method proposed by Zhang \textit{et al.} \cite{zhang2021interpatient} adversarial CNN in sensitivity for VEB and F1-score and outperformed CWT-based 2D CNN proposed by Wang \textit{et al.} \cite{wang2021automatic} in sensitivity for SVEB.  According to the author observation, the proposed model presented higher sensitivity for minority classes (SVEB and VEB) and a much lower computational complexity, with only 1,267–1,619 parameters compared to the significantly larger architectures of competing models. 
        
        For edge deployment, the model was extensively optimized using TensorFlow Lite (TFLite) tools. Weight quantization and pruning were applied to minimize the model size and computational demands. The final quantized model had a size of only 15 KB and achieved an average inference time of less than 1 ms on a Raspberry Pi device. The model’s memory usage ranged between 12 MB and 24 MB, depending on the level of quantization and pruning applied. 

    
        Raj and Ray \cite{raj2018personalized} presented a methodology for real-time ECG classification and arrhythmia detection, optimized for deployment on resource-constrained hardware. The study combined the Discrete Orthogonal Stockwell Transform (DOST) for feature extraction and an Artificial Bee Colony (ABC)-optimized Least-Square Twin Support Vector Machine (LSTSVM) for classification, aiming to achieve high accuracy while addressing challenges posed by inter-patient variability and hardware limitations.
        
        The methodology began with preprocessing ECG signals to improve signal quality by removing baseline wander, high-frequency noise, and other artifacts using median filters and a low-pass filter. The signals were segmented based on R-peak detection using the Pan-Tompkins algorithm, with each segment consisting of 256 samples centered around the R-peak. Feature extraction was performed using DOST, which provided time-frequency coefficients that captured morphological and temporal variations in the ECG signals. These coefficients effectively represented the complex dynamics of arrhythmic heartbeats. For classification, the extracted features served as the input of an ABC-optimized LSTSVM classifier. The ABC algorithm optimized the classifier’s hyperparameters, such as penalty terms and kernel parameters, to improve performance and computational efficiency. The system utilized a radial basis function (RBF) kernel for nonlinear classification, while a directed acyclic graph (DAG) technique was employed to handle multi-class classification efficiently.
        
        The study was validated using the MIT-BIH Arrhythmia Database under two evaluation schemes. The class-oriented scheme included all 16 arrhythmia classes in the dataset and used random partitions of data for training and testing. The personalized scheme followed the inter-patient paradigm, splitting training and testing sets by patient records to simulate real-world scenarios. Two setups were evaluated: Setup IIA, where 20 records were used for training and 24 for testing, and Setup IIB, where 22 records were allocated to training and 22 to testing. A hardware prototype was implemented on an ARM9 microcontroller to demonstrate the feasibility of the system for real-time ECG monitoring. The prototype performed signal acquisition, preprocessing, feature extraction, and classification in real time, with results displayed on a 16×2 LCD.
        
        In terms of performance, the system achieved an accuracy of 96.29\% in the class-oriented scheme, with high sensitivity and precision across most classes. In the personalized scheme, Setup IIA reported an accuracy of 96.08\%, demonstrating strong generalization to unseen patient data, while Setup IIB achieved an accuracy of 86.89\%, reflecting the challenges of inter-patient variability. 
    
        Xing \textit{et al.} \cite{xing2022accurate} proposed a lightweight ECG classification model that combined a Spiking Neural Network (SNN) with a Channel-Wise Attention Mechanism (CAM). Their method began with ECG signal denoising using wavelet transform thresholding. The Daubechies 5 (db5) wavelet was selected, and a soft-thresholding approach was applied to remove noise, such as baseline wander and power-line interference, while preserving signal morphology. Segmentation of the ECG signal was performed using annotations from the MIT-BIH Arrhythmia Database. Each segment consisted of 300 samples centered on the R-peak, capturing the P, QRS, and T waves. The model processed these segments using a spiking neural network with Leaky Integrate-and-Fire (LIF) neurons, which encoded the ECG data into sparse binary spikes. 
        
        To enhance the model's ability to extract relevant features, a Channel-Wise Attention Mechanism (CAM) was integrated into the SNN architecture. The CAM assigned importance weights to different channels of the ECG data, enabling the model to focus on discriminative features.
        
        The study evaluated the model under three strategies: the ``70:30'' strategy (random division of 70\% training and 30\% testing), the inter-patient strategy (strict separation by patients), and the patient-specific strategy (training on part of a patient's data and testing on the remainder). Following De Chazal inter-patient approach \cite{de_chazal_automatic_2004}, 22 records were used for training and the remaining 22 for testing, ensuring the model's robustness to inter-patient variability.
        
        The model was deployed on an Artix-7 FPGA. The deployment achieved an inference time of 1.37 ms per beat under the ``70:30'' strategy and 1.32 ms per beat for the inter-patient and patient-specific strategies. The energy consumption was 346.33 $\mu$J per beat (70:30 strategy), 246 mW (inter-patient strategy), and 324.51 $\mu$J per beat (patient-specific strategy).
        
        In terms of classification performance, the model achieved an overall accuracy of 98.26\% under the ``70:30'' strategy, 92.07\% under the inter-patient strategy, and 98.2\% for the patient-specific strategy. Sensitivity and F1-scores were also high, with the F1-score reaching 89.09\% and 90.9\% for the ``70:30'' and patient-specific strategies, respectively. The CAM significantly improved feature extraction, as demonstrated by ablation studies, particularly in handling challenging arrhythmias such as ventricular ectopic beats. 
    
        Mao \textit{et al.} \cite{mao2022ultra} proposed an ECG classification processor that integrates Spiking Neural Networks (SNN) and Artificial Neural Networks (ANN) 
        in a reconfigurable architecture to achieve ultra-low energy consumption and high classification accuracy. The processor leveraged the high training accuracy of ANNs and the energy-efficient inference of SNNs by converting pre-trained ANN weights into SNN weights using a threshold balancing and weight rescaling method. This dual-purpose design enabled the system to switch between ANN training for personalized calibration and SNN inference for real-time classification, optimizing both performance and power efficiency.
    
        The methodology incorporated a dual-purpose binary encoding scheme to reduce computational complexity and memory usage. ECG signals were encoded into 96-bit binary representations, significantly reducing the size compared to the typical 400 samples per heartbeat. This encoding scheme achieved a 3.66x reduction in computational complexity and 50x savings in storage size, while maintaining high accuracy. 
        
        It achieved an initial classification accuracy of 93.67\%, which improved to 97.36\% after on-chip ANN learning for patient-specific calibration. Specificity increased from 55.54\% to 82.30\%, and the F1-score improved from 96.78\% to 98.66\%, demonstrating significant gains in sensitivity and precision after calibration. The energy efficiency of the processor was great, with an energy consumption of only 0.3 $\mu\text{J}$ per heartbeat during SNN inference and 1.19 $\mu\text{J}$ per heartbeat during ANN on-chip learning.    

    \subsection{Comparative Analysis of ECG Classification Methods}

        Analyzing these five studies is essential for identifying state-of-the-art ECG classification methods, as they address key challenges like class imbalance, minority class accuracy, and optimization for resource-constrained devices. They highlight clinically relevant approaches that balance accuracy, efficiency, and real-time feasibility, advancing ECG systems for clinical and wearable healthcare applications. In this context, Table \ref{tab:comparison} provides a detailed summary of the results from the five studies that meet all the specified criteria, including method descriptions, overall accuracy, inference time, energy consumption, model size, and class-specific performance metrics (sensitivity, positive predictive value, and F1-score) for arrhythmia detection across N, S, V, F, and Q classes. 
    
        \begin{table}[!htb]
            \centering
            \caption{Comparison of ECG classification methods based on key metrics, including model average accuracy (Acc\_avg), inference time (Time), storage size (Size), energy consumption (EC), and class-specific performance (N, S, V, F, Q) across different state-of-the-art studies.}
            \label{tab:comparison}
            \scriptsize
            \begin{tabular}{|c|c|>{\centering\arraybackslash}m{2.8cm}|>{\centering\arraybackslash}m{2.5cm}|>{\centering\arraybackslash}m{2.5cm}|>{\centering\arraybackslash}m{2.5cm}|>{\centering\arraybackslash}m{2.5cm}|}
                \hline
                \multicolumn{2}{|c|}{\textbf{Article}} & \textbf{Yan \textit{et al.} \cite{yan2021energy}} & \textbf{Farag \cite{farag2023tiny}} & \textbf{Raj and Ray \cite{raj2018personalized}} & \textbf{Xing \textit{et al.} \cite{xing2022accurate}} & \textbf{Mao \textit{et al.} \cite{mao2022ultra}} \\ \hline
                \multicolumn{2}{|c|}{\textbf{Method}} & Two-stage CNN with early stopping, converted to SNN for inference. & Matched Filter-based 1D CNN with RR intervals, optimized for edge deployment with quantization and pruning. & Discrete Orthogonal Stockwell Transform (DOST) for feature extraction, combined with ABC-optimized Least-Square Twin SVM (LSTSVM). & Spiking Neural Network (SNN) integrated with Channel Attentional Mechanism (CAM), optimized for real-time ECG classification. & ANN-to-SNN conversion with on-chip ANN learning for patient-specific calibration. \\ \hline
                \multirow{4}{*}{\textbf{N}} & \textbf{Acc} & - & - & - & - & - \\ \cline{2-7} 
                 & \textbf{Sen} & 92\% & 96.96\% / \textbf{99.10}\% & 88.50\% & 98.37\% & - \\ \cline{2-7} 
                 & \textbf{Pp} & 97\% & \textbf{99.38\%} / 99\% & 98.54\% & 93.23\% & - \\ \cline{2-7} 
                 & \textbf{F1-S} & - & \textbf{98.16\%} / 99.05\% & 93.25\% & 95.73\% & - \\ \hline
                \multirow{4}{*}{\textbf{S}} & \textbf{Acc} & - & - & - & - & - \\ \cline{2-7} 
                 & \textbf{Sen} & 67\% &\textbf{ 85.30\%} / 81.60\% & 72.29\% & - & - \\ \cline{2-7} 
                 & \textbf{Pp} & 46\% & 58.06\% / \textbf{82.68\%} & 52.06\% & - & - \\ \cline{2-7} 
                 & \textbf{F1-S} & - & 69.09\% / \textbf{82.14\%} & 60.53\% & - & - \\ \hline
                \multirow{4}{*}{\textbf{V}} & \textbf{Acc} & - & - & - & - & - \\ \cline{2-7} 
                 & \textbf{Sen} & 77\% & \textbf{96.34\%} / 95\% & 81.59\% & 69.04\% & - \\ \cline{2-7} 
                 & \textbf{Pp} & 59\% & 90.20\% / \textbf{95.63\%} & 62.35\% & 76.46\% & - \\ \cline{2-7} 
                 & \textbf{F1-S} & - & 93.17\% / \textbf{95.31\%} & 70.68\% & 72.56\% & - \\ \hline
                \multirow{4}{*}{\textbf{F}} & \textbf{Acc} & - & - & - & - & - \\ \cline{2-7} 
                 & \textbf{Sen} & - & - & \textbf{17.78\%} & - & - \\ \cline{2-7} 
                 & \textbf{Pp} & - & - & \textbf{2.31\%} & - & - \\ \cline{2-7} 
                 & \textbf{F1-S} & - & - & \textbf{4.08\%} & - & - \\ \hline
                \multirow{4}{*}{\textbf{Q}} & \textbf{Acc} & - & - & - & - & - \\ \cline{2-7} 
                 & \textbf{Sen} & - & - & \textbf{14.28\%} & - & - \\ \cline{2-7} 
                 & \textbf{Pp} & - & - & \textbf{0.48\%} & - & - \\ \cline{2-7} 
                 & \textbf{F1-S} & - & - & \textbf{0.94\%} & - & - \\ \hline
                \multicolumn{2}{|c|}{\textbf{Acc\_avg}} & 90\% & 96.48\% / \textbf{98.18\%} & 86.89\% & 92.07\% & 97.36\% \\ \hline
                \multicolumn{2}{|c|}{\textbf{Time}} & - & \textbf{\textless{}1 ms} & - & 1.32 ms & - \\ \hline
                \multicolumn{2}{|c|}{\textbf{Size}} & - & 15 KB & - & - & \textbf{8 KB} \\ \hline
                \multicolumn{2}{|c|}{\textbf{EC}} & \textbf{77 mW} & - & - & 246 mW & 0.3 $\mu$J / beat \\ \hline
            \end{tabular}
        \end{table}
    
        According to Table \ref{tab:comparison}, the approach proposed by Yan \textit{et al.} \cite{yan2021energy} based on a two-stage CNN converted to SNN for inference, emphasizes energy efficiency, consuming only 77 mW, the lowest reported energy consumption metric. However, their method exhibits moderate performance across the other studies. While it achieved respectable sensitivity (92\%) and positive predictive value (97\%) for the N class, its results for minority classes such as S and V were lower, with sensitivities of 67\% and 77\%, respectively. This highlights limitations in handling class imbalances and complex arrhythmias. 
    
        The method proposed by Farag \textit{et al.} \cite{farag2023tiny} 
        was based on the two best-performing models, MF 1 and MF 6, which represent distinct configurations of their matched filter-based convolutional neural network (MF-CNN). MF 1 utilized a kernel size of 64, a default initialization for the convolutional layer, and a derivative input feature. This configuration focused on achieving high sensitivity for minority classes, such as the VEB class. On the other hand, MF 6 was configured with a kernel size of 32, a trainable convolutional layer initialized with matched filter templates, and no class weights. This setup emphasized balancing precision and sensitivity, yielding superior overall accuracy and F1-scores, which can be seen in the reported metrics of Table \ref{tab:comparison}, where the left values are from MF 1 and the right ones from MF 6.
     
        Their method stands out for its consistently high performance across most metrics. For the majority class (N), it achieved excellent sensitivity (96.96\% / 99.10\%), positive predictive value (99.38\%), and F1-score (98.16\% / 99.05\%), outperforming other methods. Similarly, for the S class, it recorded the highest sensitivity (85.30\%) and F1-score (82.14\%), reflecting its robustness in detecting these arrhythmias. The model's compact size (15 KB) and ultra-fast inference time ($<1$ ms) make it particularly suitable for edge deployment in real-time scenarios. However, the lack of reported results for minority classes such as fusion beats (F) and unclassifiable beats (Q) leaves gaps in its evaluation for comprehensive arrhythmia detection.   
        
        Raj and Ray \cite{raj2018personalized} proposed a method based on Discrete Orthogonal Stockwell Transform (DOST) and Least-Square Twin SVM, which is the only method reporting performance for minority classes F and Q. However, its sensitivity for these classes was low (17.78\% for F and 14.28\% for Q), with corresponding F1-scores of 4.08\% and 0.94\%, reflecting significant difficulties in handling class imbalances. The poor positive predictive value for these classes (2.31\% for F and 0.48\% for Q) further indicates a high rate of false positives. While it achieved decent performance for the N class (88.50\% sensitivity, 98.54\% positive predictivity), its overall accuracy (86.89\%) was the lowest among the methods, suggesting limited generalization to unseen patient data.
    
        The approach presented by Xing \textit{et al.} \cite{xing2022accurate} integrates a channel-wise attentional mechanism (CAM) with an SNN, and showed strong performance for the N class with one of the highest sensitivity (98.37\%) among all methods. However, its sensitivity for V (69.04\%) and the lack of reported metrics for S, F, and Q classes indicate challenges in generalizing to more complex arrhythmias. Additionally, its energy consumption (246 mW) and inference time (1.32 ms) suggest higher computational demands compared to the method presented by Farag \textit{et al.} \cite{farag2023tiny}, which achieved better efficiency.
    
        Finally, the method proposed by Mao \textit{et al.} \cite{mao2022ultra} achieved one of the highest overall accuracy (97.36\%) while demonstrating exceptional energy efficiency (0.3 $\mu$J per heartbeat) and the most compact reported model size (8 KB). This makes it ideal for wearable devices and portable healthcare systems. However, the method lacks specific results for individual ECG classes, particularly the minority ones, limiting its ability to be critically compared in terms of class-specific performance. While the ANN-to-SNN conversion and on-chip learning significantly enhance its efficiency, more detailed reporting of metrics across all classes would provide a better understanding of its robustness.
    
        Besides that, Mao \textit{et al.} \cite{mao2022ultra} highlighted the improvements in classification accuracy achieved through on-chip learning compared to off-chip learning across eight patient records. The average accuracy increased from 93.67\% (off-chip) to 97.36\% (on-chip), with gains in patient-specific data adaptation. On-chip learning fine-tuned the model for individual patients, particularly improving performance in challenging arrhythmias like VEB beats, where off-chip methods struggled.

        According to Table \ref{tab:comparison} and the subsequent discussion, the study by Farag \textit{et al.} \cite{farag2023tiny} achieved the highest performance across the three main ECG classes (N, S, V) and recorded the highest average accuracy. Additionally, it reported the lowest inference time among the five studies analyzed. The only aspect where it fell short was storage size, which was larger compared to Mao \textit{et al.} \cite{mao2022ultra}. However, the lack of class-specific performance metrics in Mao’s study makes a direct comparison between the two methods difficult. Given these findings, Farag \textit{et al.} \cite{farag2023tiny} stands out as the state-of-the-art, in terms of metrics, among methods that satisfy all criteria essential for real-world ECG classification and deployment in resource-constrained environments.

        Although Farag \textit{et al.} \cite{farag2023tiny} achieved superior classification performance, their model was trained externally before being deployed on an edge device for inference. Their approach leveraged quantization and pruning to optimize the model for real-time execution on a Raspberry Pi, achieving an inference time of less than 1 ms. In contrast, Mao \textit{et al.} \cite{mao2022ultra} is the only work, among the five selected articles, that implemented on-chip learning, meaning the model was trained directly on the hardware, allowing for patient-specific adaptation, improved classification accuracy and a challenging learning environment. This approach led to an accuracy increase from 93.67\% to 97.36\% after on-chip learning while maintaining ultra-low energy consumption of only 0.3 $\mu$J per heartbeat, as mentioned before. While Farag’s externally trained model delivered better performance in key metrics, Mao’s method ensures continuous learning and adaptation within embedded systems, making it more suitable for dynamic, real-world applications where models need to adjust to patient-specific variations.

        It is important to highlight that among the analyzed studies, Yan \textit{et al.} \cite{yan2021energy}, Xing \textit{et al.} \cite{xing2022accurate}, and Mao \textit{et al.} \cite{mao2022ultra} implemented Spiking Neural Networks (SNNs) to enhance energy efficiency in ECG classification. Yan \textit{et al.} \cite{yan2021energy} trained a CNN externally and later converted it into an SNN for deployment on a Shenjing neuromorphic accelerator. Xing \textit{et al.} \cite{xing2022accurate} incorporated an SNN with a Channel-Wise Attentional Mechanism (CAM), optimizing it for real-time inference on an Artix-7 FPGA. With SNNs present in 60\% of the studies, this method emerges as a promising approach for ECG classification in resource-constrained environments.

        However, Mao \textit{et al.} \cite{mao2022ultra} and Farag \textit{et al.} \cite{farag2023tiny} explored different strategies. Mao's approach adopted an on-chip learning approach, enabling continuous adaptation directly on hardware, making it particularly suited for real-world applications where models need to adjust dynamically to patient-specific variations. In contrast, Farag's method did not employ SNNs but instead relied on a matched filter-based CNN, optimized through quantization and pruning, achieving the highest classification performance for the core ECG classes (N, S, V) while maintaining ultra-low latency for edge deployment.

    \subsection{Standardizing Performance Reporting for Real-World ECG Classification}
        
        This systematic review aims to establish a reporting guideline for ECG classification studies that seek to address real-world challenges by meeting the E3C: adherence to AAMI standards \cite{aami:2008}, application of the inter-patient method, and conducting a feasibility study for embedded implementation. To ensure meaningful and standardized comparisons between models, studies should report accuracy, positive prediction, and sensitivity for all five AAMI-recommended classes (N, S, V, F, Q), ensuring compliance with clinical classification standards. Additionally, the F1-score should be included to provide a more balanced evaluation of performance across all classes. 
        
        Beyond classification metrics, models must also report their overall accuracy, memory and energy consumption, model size, and inference time, as these factors are crucial for assessing the feasibility of deployment in real-world embedded and wearable ECG monitoring systems. By standardizing these reporting practices, this systematic review aims to facilitate fair and transparent comparisons, helping to advance the development of ECG classification methods that are both accurate and practical for real-time healthcare applications.

\section{Conclusions}
\label{sec:conclusions}

    This systematic review underscores the importance of adopting rigorous evaluation standards in ECG arrhythmia classification to ensure the clinical reliability and real-world feasibility of machine learning models. Our findings reveal that only a small fraction of the surveyed studies simultaneously adhere to AAMI recommendations, follow the inter-patient paradigm, employ the MIT-BIH Arrhythmia Database, and assess model feasibility for embedded deployment. Among these, the study by Farag \textit{et al.} \cite{farag2023tiny} emerges as the state-of-the-art, achieving superior classification performance, low inference time, and high efficiency in resource-constrained settings,  although Mao \textit{et al.} \cite{mao2022ultra} was the only approach, that satisfies all criteria, that implemented on-chip learning.
    Thus, their method enables patient-specific adaptation while creating a challenging learning environment. However, the absence of standardized reporting across studies impedes direct comparisons and fair benchmarking of different approaches.

    The studies analyzed highlight gaps, including non-adherence to AAMI standards, limited adoption of the inter-patient paradigm, and insufficient research on the feasibility of deploying models on low-power hardware, which better aligns with real-world constraints. These limitations underscore key research questions: How can ECG classification models be standardized to ensure fair evaluation? What optimization strategies can enable real-time, low-power ECG classification on embedded devices?
    Addressing these challenges requires the development of benchmarking frameworks aligned with AAMI standards, as well as the exploration of efficient model optimization techniques, including quantization, pruning, lightweight architectures, FPGA acceleration, and on-chip learning, which have demonstrated their effectiveness in embedded deployment, as highlighted in this work.

    Addressing these challenges requires not only methodological advancements but also a more structured approach to model evaluation and reporting. To bridge this gap, we propose a set of reporting guidelines that emphasize the inclusion of class-specific performance metrics, overall model efficiency, and computational feasibility for embedded applications. Standardizing these practices will enable a more accurate assessment of ECG classification models, fostering advancements in real-time arrhythmia detection for wearable and portable healthcare solutions. Future research should focus on developing models according to E3C, that balance high diagnostic accuracy with low computational complexity, ensuring their practical applicability in real-world healthcare environments. By tackling these critical challenges, the field can progress toward more transparent, reproducible, and impactful innovations in automated ECG classification.

\section*{Acknowledgments}
    The authors would also like to thank the \textit{Coordenação de Aperfeiçoamento de Pessoal de Nível Superior} - Brazil (CAPES) - Finance Code 001, \textit{Fundacão de Amparo à Pesquisa do Estado de Minas Gerais} (FAPEMIG, grants APQ-01518-21), \textit{Conselho Nacional de Desenvolvimento Científico e Tecnológico} (CNPq, grants 308400/2022-4) and Universidade Federal de Ouro Preto (PROPPI/UFOP) for supporting the development of this study.

\bibliographystyle{main}  
\bibliography{main}  

\end{document}